\documentclass{article}

\usepackage{microtype}
\usepackage{graphicx}
\usepackage{subfigure}
\usepackage{booktabs} % for professional tables
\usepackage[dvipsnames]{xcolor}
\usepackage{caption}
\usepackage{url}
\usepackage{amsmath, amsthm, amssymb, dsfont}
\usepackage[noend]{algpseudocode}
\usepackage{amsmath,amsfonts,bm}

\def\eqref#1{eq. ~\ref{#1}}
\def\1{\bm{1}}

\DeclareMathAlphabet{\mathsfit}{\encodingdefault}{\sfdefault}{m}{sl}
\SetMathAlphabet{\mathsfit}{bold}{\encodingdefault}{\sfdefault}{bx}{n}

\def\gL{{\mathcal{L}}}

\def\gP{{\mathcal{P}}}

\def\gS{{\mathcal{S}}}

\def\sR{{\mathbb{R}}}

\newcommand{\bx}{\boldsymbol{x}}
\newcommand{\bv}{\boldsymbol{v}}
\newcommand{\by}{\boldsymbol{y}}

\newtheorem{theorem}{Theorem}
\newtheorem{prop}{Proposition}

\newcommand\ourmethod{pFedHN}

\usepackage[accepted, nohyperref]{icml2021}

\icmltitlerunning{Personalized Federated Learning using Hypernetworks}

\begin{document}

\twocolumn[
\icmltitle{Personalized Federated Learning using Hypernetworks}


\begin{icmlauthorlist}
\icmlauthor{Aviv Shamsian$^*$}{biu}
\icmlauthor{Aviv Navon$^*$}{biu}
\icmlauthor{Ethan Fetaya}{biu}
\icmlauthor{Gal Chechik}{biu,nvidia}
\end{icmlauthorlist}
\icmlaffiliation{biu}{Bar-Ilan University, Ramat Gan, Israel}
\icmlaffiliation{nvidia}{Nvidia, Tel-Aviv, Israel}

\icmlcorrespondingauthor{Aviv Shamsian}{aviv.shamsian@live.biu.ac.il}
\icmlcorrespondingauthor{Aviv Navon}{aviv.navon@biu.ac.il}

% Academic affiliations should list Department, University, City, Region, Country % Industry affiliations should list Company, City, Region, Country

% You can specify symbols, otherwise they are numbered in order. % Ideally, you should not use this facility. Affiliations will be numbered % in order of appearance and this is the preferred way. % You may provide any keywords that you % find helpful for describing your paper; these are used to populate % the "keywords" metadata in the PDF but will not be shown in the document
\icmlkeywords{Machine Learning, ICML}

\vskip 0.3in
]

\printAffiliationsAndNotice{\icmlEqualContribution}


% This command actually creates the footnote in the first column % listing the affiliations and the copyright notice.
% The command takes one argument, which is text to display at the start of the footnote. % The \icmlEqualContribution command is standard text for equal contribution. % Remove it (just {}) if you do not need this facility.

%\printAffiliationsAndNotice{}  % leave blank if no need to mention equal contribution % \printAffiliationsAndNotice{\icmlEqualContribution} % otherwise use the standard text.

\begin{abstract}
Personalized federated learning is tasked with training machine learning models for multiple clients, each with its own data distribution. The goal is to train personalized models in a collaborative way while accounting for data disparities across clients and reducing communication costs.

We propose a novel approach to this problem using hypernetworks, termed \textit{\ourmethod{}} for \textit{personalized Federated HyperNetworks}. In this approach, a central hypernetwork model is trained to generate a set of models, one model for each client. This architecture provides effective parameter sharing across clients, while maintaining the capacity to generate unique and diverse personal models. Furthermore, since hypernetwork parameters are never transmitted, %, only the parameters of client networks,
this approach decouples the communication cost from the trainable model size. We test \ourmethod{} empirically in several personalized federated learning challenges and find that it outperforms previous methods. Finally, since hypernetworks share information across clients we show that \ourmethod{} can generalize better to new clients whose distributions differ from any client observed during training.

%Personalized federated learning is tasked with training machine learning models where the data is distributed between clients while taking into account the data disparity between these clients. The goal is to improve performance by jointly training across clients without forcing all clients to use a singular global model. We propose a new approach to handle this problem by using hypernetworks to generate the unique models per client. This allows us to    

\end{abstract}

\section{Introduction}
\label{intro}
Federated learning (FL) is the task of learning a model over multiple disjoint local datasets \cite{McMahan2017CommunicationEfficientLO, Yang2019FederatedML}. It is particularly useful when local data cannot be shared due to privacy, storage, or communication constraints. This is the case, for instance, in IoT applications that create large amounts of data at edge devices, or with medical data that cannot be shared due to privacy \citep{Wu2020PersonalizedFL}. In federated learning, all clients collectively train a shared model without sharing data and while trying to minimize communication.
Unfortunately, learning a single global model may fail when the data distribution varies across clients. For example, user data may come from different devices or geographical locales and is potentially heterogeneous. In the extreme, each client may be required to solve a different task. To handle such heterogeneity across clients, \textit{Personalized Federated Learning} (PFL) \cite{smith2017federated} allows each client to use a \textit{personalized} model instead of a shared global model. The key challenge in PFL is to benefit from joint training while allowing each client to keep its own unique model and at the same time limit the communication cost. While several approaches were recently proposed for this challenge, these problems are far from being resolved.  

In this work, we describe a new approach that aims to resolve these concerns. Our approach, which we name, \textit{\ourmethod{}} for \textit{personalized Federated HyperNetwork} addresses this by using hypernetworks \cite{Ha2017HyperNetworks}, a model that for each input produces parameters for a neural network. Using a single joint hypernetwork to generate all separate models allows us to perform smart parameter sharing. Each client has a unique embedding vector, which is passed as input to the hypernetwork to produce its personalized model weights. As the vast majority of parameters belong to the hypernetwork, most parameters are shared across clients. Despite that, by using a hypernetwork, we can achieve great flexibility and diversity between the models of each client. Intuitively, as the hypernetwork maps between the embedding space and the personal networks' parameter space, its image can be viewed as a low-dimensional manifold in that space. Thus, we can think of the hypernetwork as the coordinate map of this manifold. Each unique client's model is restricted to lay on this manifold and is parametrized by the embedding vector.

Another benefit of using hypernetworks is that the trained parameter vector of the hypernetwork, which is generally much larger than the parameter vectors of the clients that it produces, is never transmitted. Each client only needs to receive its own network parameters to make predictions and compute gradients. Furthermore, the hypernetwork only needs to receive the gradient or update direction to optimize its own parameters.  As a result, we can train a large hypernetwork with the same communication costs as in previous models. Compared to previous parameter sharing schemes, e.g., \citet{Dinh2020PersonalizedFL, McMahan2017CommunicationEfficientLO}, hypernetworks open new options that were not directly possible before. Consider the case where each client uses a cell phone or a wearable device, each one with different computational resources. 
%The hypernetwork can be trained to produce several networks per input, each with a different computational capacity, so that a different network could be chosen for each different client.
The hypernetwork can produce several networks per input, each with a different computational capacity, allowing each client to select its appropriate network.

This paper makes the following contributions: (1) A new approach for personalized federated learning based on hypernetworks. 
(2) This approach generalizes better (a) to novel clients that differ from the ones seen during training; and (b) to clients with different computational resources, allowing clients to have different model sizes.   
(3) A new set of state-of-the-art results for the standard benchmarks in the field CIFAR10, CIFAR100, and Omniglot.
%We show empirically that our method achieves state-of-the-art results on CIFAR10, CIFAR100, and Omniglot 

The paper is organized as follows. Section \ref{sec:model} describes our model in detail. Section \ref{sec:theory} establishes some theoretical results to provide insight into our model. Section \ref{sec:experiments} shows experimentally that \ourmethod{} achieves state-of-the-art results on several datasets and learning setups. We make our source code publicly available at: \textcolor{magenta}{\url{https://github.com/AvivSham/pFedHN}}. 

%From MOCHA~\cite{smith2017federated}:
%... challenges of federated learning~\cite{konevcny2015federated,McMahan2017CommunicationEfficientLO,konevcny2016federated}: (1) Statistical Challenges: ... (2) Systems Challenges ... .

\begin{figure*}[t]
\centering
    \begin{subfigure}[]{
    \includegraphics[width=0.57\linewidth]{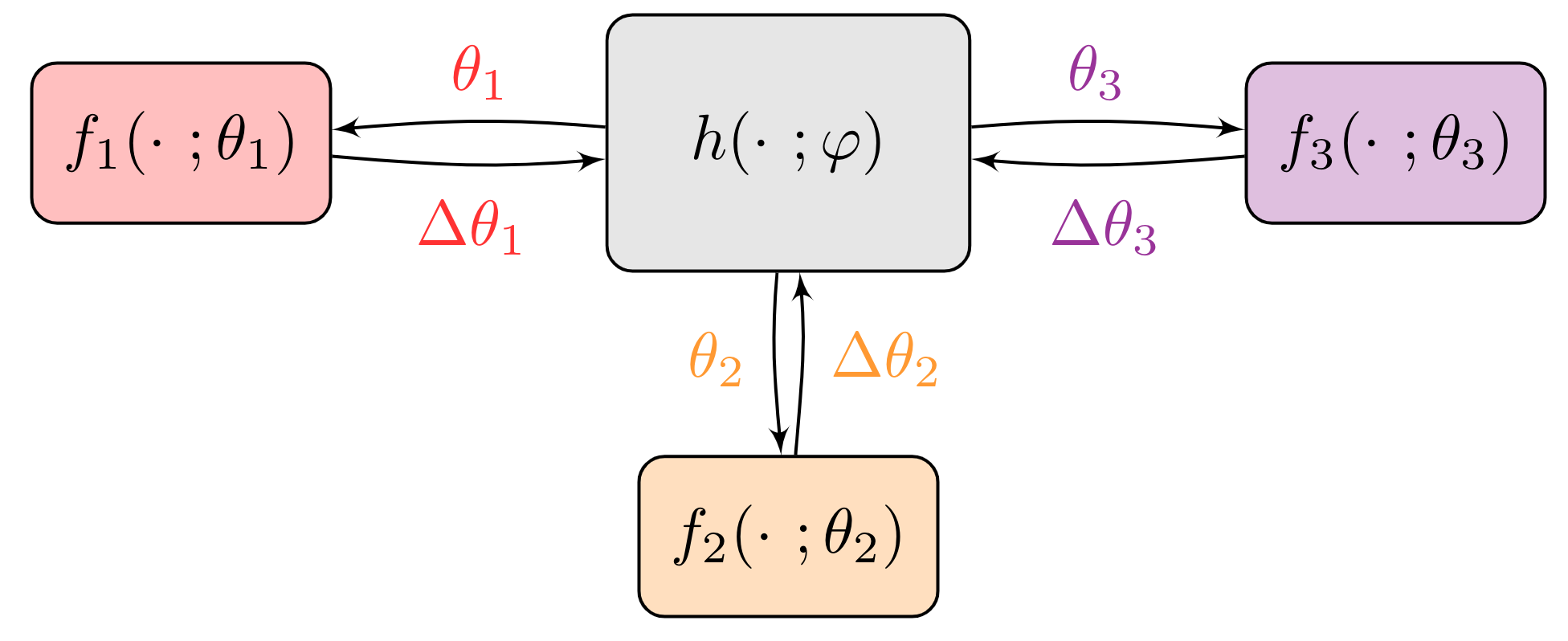}
    }
    \label{fig:global_arch}
    \end{subfigure}
    \hfill
    \begin{subfigure}[]{
    \includegraphics[width=0.37\linewidth]{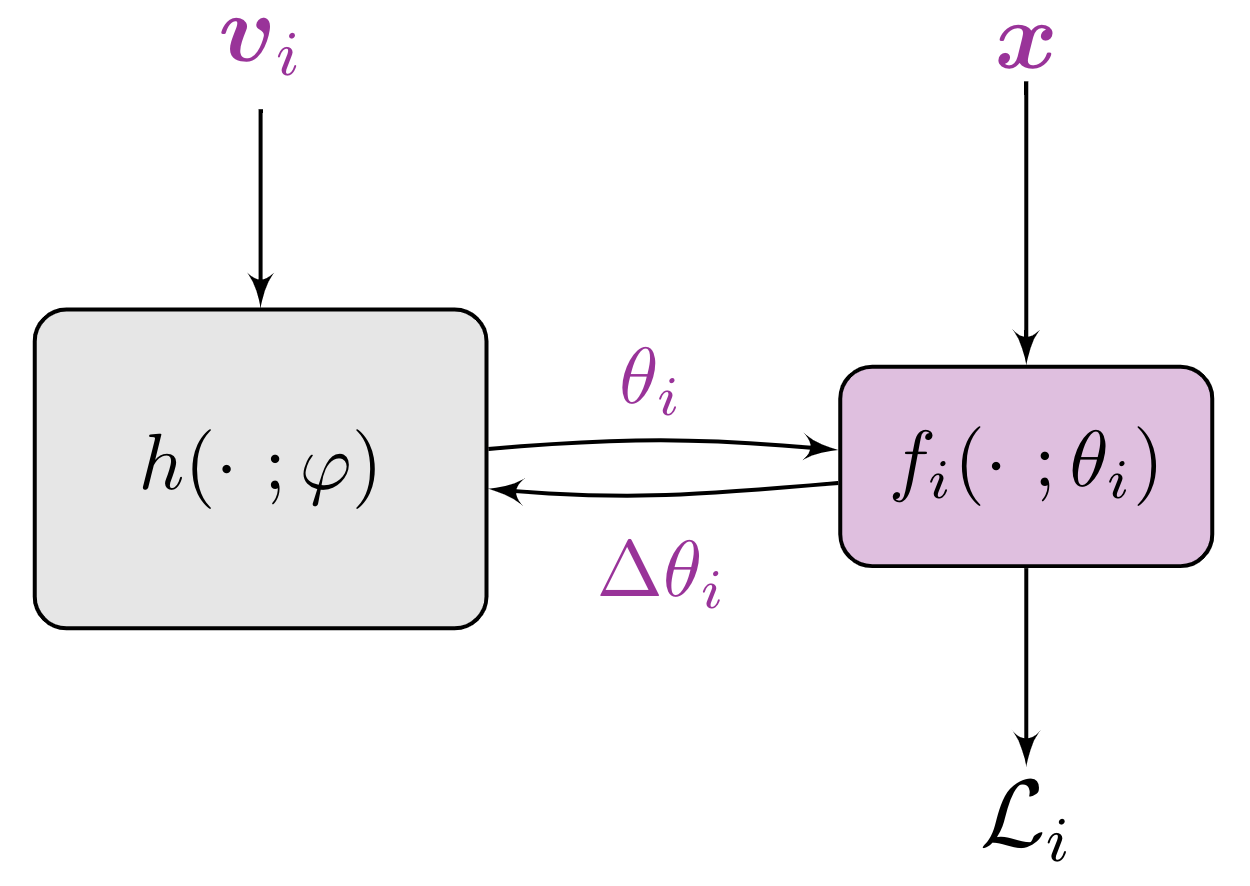}
    }
    \label{fig:arch_local}
     \end{subfigure}
    \caption{The Federated hypernetwork framework. \textbf{(a)} An HN is located on the server and communicate personal model for each clients. In turn, the clients send back the update direction $\Delta \theta_i$;
    \textbf{(b)} The HN acts on the client embedding $\bv_i$ to produce model weights $\theta_i$. The client performs several local optimization steps to obtain $\tilde{\theta}_i$, and sends back the update direction $\Delta\theta_i=\tilde{\theta}_i-\theta_i$.}
    \label{fig:arch}
\end{figure*}

\section{Related Work}

\label{related}
\subsection{Federated Learning}
 Federated learning (FL)~\citep{McMahan2017CommunicationEfficientLO, kairouz2019advances, mothukuri2021survey, Li2019FederatedLS, li2020federated} is a learning setup in machine learning in which multiple clients collaborate to solve a learning task while maintaining privacy and communication efficiency. Recently, numerous methods have been introduced for solving the various FL challenges. \citet{duchi2014privacy, mcmahan2017learning, agarwal2018cpsgd, zhu2019federated} proposed new methods for preserving privacy, and \citet{reisizadeh2020fedpaq, dai2019hyper, basu2020qsparse, li2020acceleration, stich2018local} focused on reducing communication cost. 
While some methods assume a homogeneous setup, in which all clients share a common data distribution~\citep{wang2018cooperative, lin2018don}, others tackle the more challenging heterogeneous setup in which each client is equipped with its own data distribution~\citep{zhou2017convergence, hanzely2020federated, zhao2018federated, sahu2018convergence, karimireddy2019scaffold, haddadpour2019convergence,Hsu2019MeasuringTE}.

Perhaps the most known and commonly used FL algorithm is FedAvg~\citep{McMahan2017CommunicationEfficientLO}. It learns a global model by aggregating local models trained on IID data. However, the above methods learn a shared global model for all clients instead of personalized per-client solutions.

\subsection{Personalized Federated Learning}

The federated learning setup presents numerous challenges including data heterogeneity (differences in data distribution), device heterogeneity (in terms of computation capabilities, network connection, etc.), and communication efficiency~\citep{kairouz2019advances}.
Especially data heterogeneity makes it hard to learn a single shared global model that applies to all clients. To overcome these issues, Personalized Federated Learning (PFL) aims to personalize the global model for each client in the federation~\citep{Kulkarni2020SurveyOP}.
Many papers proposed a decentralized version of the model agnostic meta-learning (MAML) problem 
%to achieve better performance proved by empirical experiments and theoretical guarantees
~\citep{Fallah2020PersonalizedFL, li2017meta, behl2019alpha, zhou2019efficient, fallah2020convergence}. Since MAML approach relies on the Hessian matrix, which is computationally costly, several works attempted to approximate the Hessian \citep{finn2017model, nichol2018first}. Another approach to PFL is model mixing where the clients learn a mixture of the global and local models \citep{deng2020adaptive, arivazhagan2019federated}.
\citet{hanzely2020federated} introduced a new neural network architecture that is divided into base and personalized layers. The central model trains the base layers by FedAvg and the personalized layers (also called top layers) are trained locally.
\citet{liang2020think} presented LG-FedAvg a mixing model where each client obtains local feature extractor and global output layers. This is an opposite approach to the conventional mixing model that enables lower communication costs as the global model requires fewer parameters. Other approaches to train the global and local models under different regularization \citep{Huang2020PersonalizedCF}. \citet{Dinh2020PersonalizedFL} introduced pFedMe, a method that uses Moreau envelops as the client regularized loss. This regularization helps to decouple the personalized and global model optimizations.  
Alternatively, clustering methods for federated learning assume that the local data of each client is partitioned by nature \citep{Mansour2020ThreeAF}. Their goal is to group together similar clients and train a centralized model per group. In case of heterogeneous setup, some clients are "closer" than others in terms of data distribution. Based on this assumption and inspired by FedAvg, \citet{zhang2020personalized} proposed pFedFOMO, an aggregation method where each client only federates with a subset of relevant clients.

\subsection{Hypernetworks}

%  Early works proposed the idea of one network that predicts the weights of another \citep{Klein2015ADC, Riegler2015ConditionedRM}, later \citet{Ha2017HyperNetworks} named it as Hypernetworks (HN). HN are deep neural-networks that output the weights for another network. In recent years, HN are widely used in various machine learning domains \citep{Brock2018SMASHOM, Lorraine2018StochasticHO, Oswald2019COL}. \citet{Littwin2019DeepMF} utilized HN for 3D shape reconstruction which leads to more accurate shape inference from a 2D projection. HN can be efficiently used for decoding error-correcting codes \citep{DBLP:conf/nips/NachmaniW19}.
%  \citet{Navon2020LearningTP} recently proposed a HN based method to find continuous set of personalized models on various MTL tasks.

Hypernetworks (HNs)~\citep{Klein2015ADC, Riegler2015ConditionedRM, Ha2017HyperNetworks} are deep neural networks that output the weights of another target network, that performs the learning task. The idea is that the output weights vary depending on the input to the hypernetwork.

HNs are widely used in various machine learning domains, including language modeling~\citep{suarez2017character}, computer vision~\citep{Ha2017HyperNetworks, klocek2019hypernetwork}, continual learning~\citep{von2019continual}, hyperparameter optimization~\citep{Lorraine2018StochasticHO, MacKay2019SelfTuningNB, Bae2020DeltaSTNEB}, multi-objective optimization~\citep{navon2021learning}, and decoding block codes~\citep{DBLP:conf/nips/NachmaniW19}. 

HNs are naturally suitable for learning a diverse set of personalized models, as HNs dynamically generate target networks conditioned on the input.

 \section{Method}\label{sec:model}
 
%  \begin{figure*}[t]
% \centering
%     \begin{subfigure}[]{
%     \includegraphics[width=0.55\linewidth]{figures/arch.png}
%     }
%     \label{fig:arch}
%     \end{subfigure}
%     \hfill
%     \begin{subfigure}[]{
%     \includegraphics[width=0.35\linewidth]{figures/arch_local.png}
%     }
%     \label{fig:arch_local}
%      \end{subfigure}
%     \caption{\gal{Current figure 1 highlights what is communicated from server to clients. 
% I would also like to have a figure that highlights the client embedding point of view. It is more central to our story, and does not appear in current Fig 1} \an{WDYM? any suggestions?}
%     The Federated hypernetwork framework. \textbf{(a)} An HN is located on the server and communicate personal model for each clients. In turn, the clients send back the update direction $\Delta \theta_i$;
%     \textbf{(b)} The HN acts on the client embedding $\bv_i$ to produce model weights $\theta_i$. The client performs several local optimization steps to obtain $\tilde{\theta}_i$, and sends back the update direction $\Delta\theta_i=\tilde{\theta}_i-\theta_i$.}
%     \label{fig:arch}
% \end{figure*}

 In this section, we first formalize the personalized federated learning (PFL) problem, then we present our \textit{personalized Federated HyperNetworks} (\ourmethod{}) approach.
 
 % ------------------------------
\subsection{Problem Formulation}
Personalized federated learning (PFL) aims to collaboratively train personalized models for a set of $n$ clients, each with its own personal private data. Unlike conventional FL, each client $i$ is equipped with its own data
distribution $\gP_i$ on $\mathcal{X}\times\mathcal{Y}$. Assume each client has access to $m_i$ IID samples from $\gP_i$, $\gS_i=\{(\bx_j^{(i)}, y_j^{(i)})\}_{i=1}^{m_i}$. Let $\ell_i:\mathcal{Y}\times\mathcal{Y}\to \mathbb{R}_+$ denote the loss function corresponds to client $i$, and $\gL_i$ the average loss over the personal training data $\gL_i(\theta_i)=\frac{1}{m_i}\sum_j \ell_i(\bx_j,y_j; \theta_i)$. Here $\theta_i$ denotes the personal model of client $i$. The PFL goal is to optimize 
\begin{equation}
    \Theta^* =\arg\min_{\Theta}\frac{1}{n}\sum_{i=1}^n\mathbb{E}_{\bx,y\sim\gP_i}[\ell_i(\bx_j,y_j; \theta_i)]    
\end{equation}
and the training objective is given by 
\begin{align}\label{pfl_problem}
    \arg\min_{\Theta} \frac{1}{n} \sum_{i=1}^n \gL_i(\theta_i)= \arg\min_{\Theta}\frac{1}{n}\sum_{i=1}^n\frac{1}{m_i} \sum_{j=1}^{m_i} \ell_i(\bx_j,y_j; \theta_i)
\end{align}
where $\Theta$ denotes the collection of all personal model parameters $\{\theta_i\}_{i=1}^n$.

\subsection{Federated Hypernetworks}
\label{fhn}

% \begin{algorithm}[th]
%     \caption{Federated hypernetwork (FHN)}\small\label{alg:fhn}
%     \begin{algorithmic}[H]
%     \State \textbf{input:} $R$ --- number of rounds, $K$ --- number of local rounds, $\alpha$ --- learning rate, $\eta$ --- client learning rate
%     \For{$r=1,...,R$}
%     \State sample client $i\in [n]$
%     \State set $\theta_i=h(\bv_i;\varphi)$ and  $\tilde{\theta}_i=\theta_i$
%     \For{$k=1,...,K$}
%     \State sample mini-batch $B \subset \gS_i$ 
%     \State $\tilde{\theta}_i=\tilde{\theta}_i-\eta \nabla_{\tilde{\theta}_i}\gL_i(B)$
%     \EndFor
%     \State $\Delta\theta_i=\tilde{\theta}_i-\theta_i$
%     \State $\varphi=\varphi-\alpha \nabla_{\varphi}\theta_i^T \Delta \theta_i$
%     \State $\bv_i=\bv_i - \alpha \nabla_{\bv_i}\varphi^T \nabla_{\varphi}\theta_i^T \Delta \theta_i$
%     \EndFor
%     \State \textbf{return:} $\varphi$
%     \end{algorithmic}
% \end{algorithm}

In this section,  we describe our proposed \textit{personalized Federated Hypernetworks} (\ourmethod{}), a novel method for solving the PFL problem (eq. \ref{pfl_problem}) using hypernetworks.
Hypernetworks are deep neural networks that output the weights of another network, conditioning on its input. Intuitively, HNs simultaneously learn a family of target networks.
Let $h(\cdot;\varphi)$ denote the hypernetwork parametrized by $\varphi$ and $f(\cdot;\theta)$ the target network parametrized by $\theta$. 
The hypernetwork is located at the server and acts on a client descriptor $\bv_i$ (see Figure~\ref{fig:arch}). The descriptor can be a trainable embedding vector for the client or fixed, provided that a good client representation is known a-priori. Given $\bv_i$ the HN outputs the weights for the $i^{th}$ client $\theta_i=\theta_i(\varphi):=h(\bv_i;\varphi)$. Hence, the HN $h$ learns a family of personalized models $\{h(\bv_i;\varphi)\mid i\in [n]\}$. \ourmethod{} provides a natural way for sharing information across clients while maintaining the flexibility of personalized models, by sharing the parameters $\varphi$.

We adjust the PFL objective (eq. \ref{pfl_problem}) according to the above setup to obtain
\begin{align}\label{fhn_problem}
    \arg\min_{\varphi,\bv_1,...,\bv_n} \frac{1}{n} \sum_{i=1}^n \gL_i(h(\bv_i;\varphi)).
\end{align}

One crucial and attractive property of \ourmethod{} is that it decouples the size of $h$ and the communication cost. The amount of data transferred is determined by the size of the target network during the forward and backward communications, and does not depend on the size of $h$. Consequently, the hypernetwork can be arbitrarily large without impairing communication efficiency. Indeed, using the chain rule we have $\nabla_\varphi\gL_i=(\nabla_{\varphi}\theta_i)^T\nabla_{\theta_i}\gL_i$ so the client only needs to communicate $\nabla_{\theta_i}\gL_i$ back to the hypernetwork, which has the same size as the \textit{personal} network parameters $\theta_i$. 

In our work, we used a more general update rule $\Delta\varphi=(\nabla_{\varphi}\theta_i)^T\Delta\theta_i$ where $\Delta\theta_i$ is the change in the local model parameters after several local update steps. As the main limitation is the communication cost, we found it beneficial to perform several local update steps, on the client side per communication round. This aligns with prior work that highlighted the benefits of local optimization steps in terms of both convergence speed (hence communication cost) and final accuracy~\cite{McMahan2017CommunicationEfficientLO,huo2020faster}.
%Instead of sending back the update after a single update of the local model, performing local update steps reduce the communication costs significantly \as{cite}. Hence, it is preferable to perform several local updates, on the client side, per communication round. 
Given the current personalized parameters $\theta_i$, we perform several  local optimization steps on the personal data to obtain $\tilde{\theta}_i$. We then return the personal model update direction $\Delta\theta_i:=\tilde{\theta}_i-\theta_i$ therefore, the update for $\varphi$ is given by $(\nabla_{\varphi}\theta_i)^T(\tilde{\theta}_i-\theta_i)$. This update rule is inspired by  \citet{zhang2019lookahead}. Intuitively, suppose we have access to the optimal solution of the personal problem $\theta_i^*=\arg\min_{\theta_i}\gL_i$, then our update rule becomes the gradient of an approximation to the surrogate loss $\tilde{\gL}_i(\bv_i,\varphi)=\frac{1}{2}||\theta_i^* - h(\bv_i;\varphi)||^2_2$ by replacing $\theta_i^*$ with $\tilde{\theta}_i$. %the standard chain rule by replacing the loss criteria with $\gL(\cdot)=\frac{1}{2}||\theta_i^* - \cdot||^2_2$.
In Appendix B (Figure \ref{fig:local_opt}), 
%\ref{app:exp_details} 
we compare the results for a different number of local update steps and show considerable improvement over using the gradient, i.e., using a single step.

%Indeed, given $i$, the server sends $\theta_i$ and receives $\Delta \theta_i$, which sizes only depends on the target network $f$. The update for the parameters $\varphi$ obtained according to $\nabla\varphi=(\nabla_{\varphi}\theta_i)^T\Delta\theta_i$.
%While the natural choice for $\Delta\theta_i$ is simply the gradient $\nabla_{\theta_i}\gL_i$, this might be sub-optimal in terms of optimization process stability, communication efficiency, and final accuracy \gal{again, pretty vague}. 

\begin{algorithm}[t]
    \caption{Personalized Federated Hypernetwork}\small\label{alg:fhn}
    \begin{algorithmic}[H]
    \State \textbf{input:} $R$ --- number of rounds, $K$ --- number of local rounds, $\alpha$ --- learning rate, $\eta$ --- client learning rate
    \For{$r=1,...,R$}
    \State sample client $i\in [n]$
    \State set $\theta_i=h(\bv_i;\varphi)$ and  $\tilde{\theta}_i=\theta_i$
    \For{$k=1,...,K$}
    \State sample mini-batch $B \subset \gS_i$ 
    \State $\tilde{\theta}_i=\tilde{\theta}_i-\eta \nabla_{\tilde{\theta}_i}\gL_i(B)$
    \EndFor
    \State $\Delta\theta_i=\tilde{\theta}_i-\theta_i$
    \State $\varphi=\varphi-\alpha \nabla_{\varphi}\theta_i^T \Delta \theta_i$
    \State $\bv_i=\bv_i - \alpha \nabla_{\bv_i}\varphi^T \nabla_{\varphi}\theta_i^T \Delta \theta_i$
    \EndFor
    \State \textbf{return:} $\varphi$
    \end{algorithmic}
\end{algorithm}

\subsection{Personal Classifier}

% In some cases it is undesirable to learn the entire network end-to-end with a single hypernetwork. As an illustrative example, consider a case where clients differ only by the ordering of class labels in their output vectors. In this case, it should be possible to use a feature extractor part of the network that is shared across all clients, if each client has its own  classification layer, which is a permutation of the classifiers of the other clients. It can be challenging for the hypernetwork to output the right classification layer permutation per client, but the shared feature extractor part is easy. 
%A central network would not need to personalize the feature extraction layers.\ef{that is against the whole point in the paper} 
In some cases, it is undesirable to learn the entire network end-to-end with a single hypernetwork. As an illustrative example, consider a case where clients differ only by the label ordering in their output vectors. In this case, having to learn the right label ordering per client adds another unnecessary  difficulty if they were to learn the classification layer as well using the hypernetwork.

%it should be possible to share a joint feature extractor across all clients and keep a separate classification head per client.

% As another example, consider the case where each clients solves a completely separate task e.g. two clients each has a different number of classes in their datasets. It makes little sense to have the hypernetwork produce each unique tasks classification layer. In these cases it would be preferable for the hypernetwork to output the feature extraction part of the network, which contains most of the trainable parameters, and each client will simultaneously optimize the output layers on top of the extracted features. 
As another example, consider the case where each client solves an entirely separate task, similar to multitask learning, where the number of classes may differ between clients. It makes little sense to have the hypernetwork produce each unique task classification layer. 

In these cases, it would be preferable for the hypernetwork to produce the feature extraction part of the target network, which contains most of the trainable parameters, while learning a local output layer for each client. Formally, let $\omega_i$ denote the personal classifier parameters of client $i$. We modify the optimization problem (\eqref{fhn_problem}) to obtain,
\begin{align}\label{pfhn_problem}
    \arg\min_{\varphi,\bv_1,...,\bv_n,\omega_1,...,\omega_n} \frac{1}{n} \sum_{i=1}^n \gL_i(\theta_i,\omega_i),
\end{align}
where we define the feature extractor $\theta_i=h(\bv_i;\varphi)$, as before. The parameters $\varphi, \bv_1, ...,\bv_n$ are updated according to Alg.~\ref{alg:fhn}, while the personal parameters $\omega_i$ are updated locally using 
\begin{equation}
    \omega_i=\omega_i-\alpha\nabla_{\omega_i}\gL_i. \nonumber
\end{equation}

\section{Analysis}\label{sec:theory}

In this section, we theoretically analyze \ourmethod{}. First, we provide an insight regarding the solution for the  \ourmethod{} (Eq.~\ref{fhn_problem}), using a simple linear version of our hypernetwork. Next, we describe the generalization bounds of our framework.

% In this section we theoretically analyse a simple linear version of our hypernetwork and look at generalization bounds of our framework in order to get some insights into our model. 

\subsection{A Linear Model}

Consider a linear version of the hypernetwork, where both the target model and the hypernetwork are linear models, $\theta_i=W\bv_i$ with $\varphi:=W\in \sR^{d\times k}$ and $\bv_i\in\sR^k$ is the $i^{th}$ clients embedding. Let $V$ denote the $k\times n$ matrix whose columns are the clients embedding vectors $\bv_i$. We note that even for convex loss functions $\gL_i(\theta_i)$ the objective $\gL(W,V)=\sum_i\gL_i(W\bv_i)$ might not be convex in $(W,V)$ but block multi-convex. In one setting, however, we get a nice analytical solution.

\begin{prop}
Let $\{X_i,\by_i\}$ be the data for client $i$ and let the loss for linear regressor $\theta_i$ be $\gL_i(\theta_i)=\Vert X_i\theta_i-\by_i\Vert^2$. Furthermore assume for all $i$, $X_i^TX_i=I_d$. Define the empirical risk minimization (ERM) solution for client $i$ as  $\bar{\theta}_i=\arg\min_{\theta\in\sR^d}\Vert X_i\theta-\by_i\Vert^2$. The optimal $W,V$ minimizing $\sum_i \Vert X_iW\bv_i-\by_i \Vert^2$ are given by PCA on $\{\bar{\theta}_1,...,\bar{\theta}_n\}$, where $W$ is the top $k$ principle components and $\bv_i$ is the coefficients for $\bar{\theta}_i$ in these components.
\end{prop}

We provide the proof in Section A %~\ref{app:theory}
of the Appendix. The linear version of our \ourmethod{} performs dimensionality reduction by PCA, but unlike classical dimensionality reduction which is unaware of the learning task, \ourmethod{} uses multiple clients for reducing the dimensionality while preserving the optimal model as best as possible. This allows us to get solutions between the two extremes: A single shared model up to scaling ($k=1$) and each client training locally ($k\geq n$). We note that optimal reconstruction of the local models ($k \geq n$) is generally suboptimal in terms of generalization performance, as no information is shared across clients.

This dimensionality reduction can also be viewed as a denoising process. Assume a linear regression with Gaussian noise model, i.e., for all clients  $p(y|x)=\mathcal{N}(x^T\theta_i^*,\sigma^2)$ and that each client solves a maximum likelihood objective. From the central limit theorem for maximum likelihood estimators \cite{CLT_for_ML} we get that\footnote{Note that the Fisher matrix is the identity from our assumption that $X_i^TX_i=I_d$.} $\sqrt{n_i}(\bar{\theta}_i-\theta^*_i)\overset{d}{\longrightarrow}\mathcal{N}(0,I_d)$ where $\bar{\theta}_i$ is the maximum likelihood solution. This means that approximately $\bar{\theta}_i=\theta^*_i+\epsilon$ with  $\epsilon\sim\mathcal{N}(0,\sigma_i I)$, i.e., our local solutions $\bar{\theta}_i$ are a noisy version of the optimal model $\theta^*_i$ with isotropic Gaussian noise.  

We can now view the linear hypernetworks as performing denoising on $\bar{\theta}_i$, by PCA. PCA is a classic approach to denoising \cite{PCAdenoising} and is well suited for reducing isotropic noise when the energy of the original points is concentrated on a small dimensional subspace. Intuitively we think of our standard hypernetwork as a nonlinear extension of this approach, which has a similar effect by forcing the models to lay on a low-dimensional manifold.

%%%%%%%%%%%%%%%%%%%%%%
\setlength{\tabcolsep}{5pt}
\begin{table*}[t]
    % \small
    % \tiny
    \scriptsize
    \centering
    \caption{\textit{Heterogeneous data}. Test accuracy over $10, 50, 100$ clients on the CIFAR10, CIFAR100, and Omniglot datasets.}
    \vskip 0.11in
    \begin{tabular}{l c c c c c c c c c c}
    \toprule
    & \multicolumn{3}{c}{CIFAR10} & & \multicolumn{3}{c}{CIFAR100} && Omniglot\\
     \cmidrule{2-4}\cmidrule{6-8}\cmidrule{10-10}\\
     \# clients & 10 & 50 & 100  & &  10 & 50 & 100 && 50\\
    \midrule
    Local & $86.46 \pm 4.02$ & $68.11 \pm 7.39$ & $59.32 \pm 5.59$ && $58.98 \pm 1.38$ & $19.98 \pm 1.41$ & $15.12 \pm 0.58$ && $65.97 \pm 0.86$ \\
    FedAvg & $51.42 \pm 2.41$ & $47.79 \pm 4.48$ & $44.12 \pm 3.10$ && $15.96 \pm 0.55$ & $15.71 \pm 0.35$ & $14.59 \pm 0.40$ && $41.61 \pm 3.59$ \\
        \hline
    Per-FedAvg & $76.65 \pm 4.84$ & $83.03 \pm 0.25$ & $80.19 \pm 1.99$ && $50.14 \pm 1.06$ & $45.89 \pm 0.76$ & $48.28 \pm 0.70$ && $76.46 \pm 0.17$\\
    FedPer & $87.27 \pm 1.39$ &  $83.39 \pm 0.47$ & $80.99 \pm 0.71$ && $55.76 \pm 0.34$ & $48.32 \pm 1.46$ & $42.08 \pm 0.18$ &&  $69.92 \pm 3.12$\\
    pFedMe & $87.69 \pm 1.93$ &  $86.09 \pm 0.32$ & $85.23 \pm 0.58$ && $51.97 \pm 1.29$ & $49.09 \pm 1.10$ & $45.57 \pm 1.02$ &&  $ 69.98 \pm 0.28$\\
    LG-FedAvg & $89.11 \pm 2.66$ & $85.19 \pm 0.58$ & $81.49 \pm 1.56$ && $53.69 \pm 1.42$ & $53.16 \pm 2.18$ & $49.99 \pm 3.13$ && $72.99 \pm 5.00$\\
    \midrule
    \ourmethod{} (ours) & $90.83\pm 1.56$ & $88.38 \pm 0.29$ & $87.97 \pm 0.70$ &&  $65.74\pm 1.80$ & $59.48 \pm 0.67$ & $\mathbf{53.24 \pm 0.31}$ &&  
    $72.03 \pm 1.08$ \\
    \ourmethod{}-PC (ours) & $\mathbf{92.47 \pm 1.63}$ & $\mathbf{90.08 \pm 0.63}$ & $\mathbf{88.09 \pm 0.86}$ && $\mathbf{68.15 \pm 1.49} $ & $\mathbf{60.17 \pm 1.63}$ & $52.4 \pm 0.74$ &&  
    $\mathbf{81.89 \pm 0.15}$ \\
    \bottomrule
    \end{tabular}
    \label{tab:hetro}
\end{table*}

\subsection{Generalization}

We now investigate how \ourmethod{} generalizes using the approach of \citet{baxter2000model}. The common approach for multi-task learning with neural networks is to have a common feature extractor shared by all tasks and a per-task head operating on these features. This case was analyzed by \citet{baxter2000model}.
Conversely, here the per-task parameters are the inputs to the hypernetwork. Next, we provide the generalization guarantee under this setting and discuss its implications.
% and we wish to understand the effect on the generalization.

% and here we ask how does the generalization behave when the per-task head is in input to the hypernetwork instead. \gal{broken. PTAL}

Let $D_i=\left\{(\bx^{(i)}_j,y^{(i)}_j)\right\}_{j=1}^m$ be the training set for the $i^{th}$ client, generated by a distribution $P_i$. We denote by $\hat{\mathcal{L}}_D(\varphi,V)$ the empirical loss of the hypernetwork $\hat{\mathcal{L}}_D(\varphi,V)=\frac{1}{n}\sum_{i=1}^n\frac{1}{m} \sum_{j=1}^{m} \ell_i\left(\bx^{(i)}_j,y^{(i)}_j; h(\varphi,\bv_i)\right)$ and by ${\mathcal{L}}(\varphi,V)$ the expected loss  ${\mathcal{L}}(\varphi,V)=\frac{1}{n}\sum_{i=1}^n\mathbb{E}_{P_i}\left[\ell_i(\bx,y; h(\varphi,\bv_i))\right]$. 

We assume weights of the hypernetwork and the embeddings are bounded in a ball of radius $R$, in which the following three Lipschitz conditions hold:
\begin{enumerate}
    \item $|\ell_i(\bx,y,\theta_1)-\ell_i(\bx,y,\theta_2)|\leq L\Vert\theta_1-\theta_2\Vert$
    \item $\Vert h(\varphi,\bv)-h(\varphi',\bv)\Vert\leq L_h\Vert \varphi-\varphi'\Vert$
    \item $\Vert h(\varphi,\bv)-h(\varphi,\bv')\Vert\leq L_V\Vert\bv-\bv'\Vert$. \quad
\end{enumerate}

\begin{theorem}\label{theory_gen}
 Let the hypernetwork parameter space be of dimension $N$ and the embedding space be of dimension $k$. Under previously stated assumptions, there exists $M=\mathcal{O}\left(\frac{k}{\epsilon^2}\log\left(\frac{RL(L_h+L_V)}{\delta}\right)+\frac{N}{n\epsilon^2}\log\left(\frac{RL(L_h+L_V)}{\delta}\right)\right)$ such that if the number of samples per client $m$ is greater than $M$, we have with probability at least $1-\delta$ for all $\varphi,V$ that $|\mathcal{L}(\varphi,V)- \hat{\mathcal{L}}_D(\varphi,V)|\leq\epsilon.$
\end{theorem}

Theorem~\ref{theory_gen} provides insights on the parameter-sharing effect of \ourmethod{}. The first term for the number of required samples $M$ depends on the dimension of the embedding vectors; as each client corresponds to its unique embedding vector (i.e. not being shared between clients), this part is independent of the number of clients $n$. 
However, the second term depends on the size of the hypernetwork $N$, is reduced by a factor $n$, as the hypernetwork's weights are shared.

% The theorem is a special case of Theorem 4 in \citet{baxter2000model}, where we bound the covering number for our \ourmethod{} model. See Appendix~\ref{app:theory} for details. As one can see from the bound, there is a part that depends on the embedding size $k$ that is not affected by the number of clients $n$, while the size of the hypernetwork $N$ is reduced by a factor of $n$ as the weights of the hypernetwork are shared \gal{need to be more precise in the last sentence}\ef{What would make it clearer? We are simply stating there is a N/n factor}.

Additionally, the generalization is affected by the Lipschitz constant of the hypernetwork, $L_h$ (along with other Lipschitz constants), as it can affect the effective space we can reach with our embedding. In essence, this characterizes the price that we pay, in terms of generalization, for the hypernetworks flexibility. %While large $L_h$ can increase the effective space we can reach with our embedding, it also increases the amount of per-client samples needed to guarantee generalization (for a fixed $\epsilon$ and $\delta$).  
It might also open new directions to improve performance. However, our initial investigation into bounding the Lipschitz constant by adding spectral normalization \cite{spectral_normalization} did not show any significant improvement, see Appendix~C.

\section{Experiments}\label{sec:experiments}

% \begin{table*}[t]
%     \vskip 0.15in
%     % \small
%     \scriptsize
%     \centering
%     \caption{\textit{Computational budget}. S,M,L denote the local model size.}
%     \begin{tabular}{l c c c c c c c c}
%     \toprule
%     & \multicolumn{3}{c}{CIFAR10} && \multicolumn{3}{c}{CIFAR100} \\
%      \cmidrule{2-4} \cmidrule{6-8}\\
%     Local model size & S  & M & L && S  & M & L\\
%     \midrule
%     FedAvg & $36.91 \pm 2.26$ & $42.09 \pm 2.37$ & $47.51\pm 1.97$ && $9.79 \pm 0.19$ & $11.76 \pm 1.14$ & $17.86 \pm 2.42$\\
%     Per-FedAvg & $70.72 \pm 2.57$ & $72.33 \pm 2.57$ & $75.13 \pm 0.61$ && $41.49 \pm 0.91$ & $43.22 \pm 0.16$ & $44.03 \pm 1.77$ \\
%     pFedMe & $81.21 \pm 1.23$ & $84.08 \pm 1.63$ & $83.15 \pm 2.45$   && $39.91 \pm 0.81$ & $41.99 \pm 0.55$ & $44.93 \pm 1.63$ \\
%     LG-FedAvg & $73.93 \pm 3.65$ & $53.13 \pm 3.49$ & $54.72 \pm 2.50$ && $33.48 \pm 4.83$ & $29.15 \pm 1.51$ & $23.01 \pm 1.41$\\
%     \midrule
%     \ourmethod{} (ours) & $\mathbf{85.38 \pm 1.21}$ & $\mathbf{86.92 \pm 1.35}$ & $\mathbf{87.20 \pm 0.76}$ && $\mathbf{48.04 \pm 0.89}$ & $\mathbf{48.66 \pm 1.21}$ & $\mathbf{50.66 \pm 2.70}$\\
%     %\midrule
%     % \ourmethod{} (ours S) & $87.80 \pm 0.1$ & $-$ & $-$ && $52.24 \pm 0.71$ & $-$ & $-$ \\
%     % \ourmethod{} (ours M) & $-$ & $ 88.83 \pm 0.14 $ & $-$ && $-$ & $55.48 \pm 0.90$ & $-$ \\
%     %\ourmethod{} (ours L) & $-$ & $-$ & $89.06 \pm 0.14$ && $-$ & $-$ & $56.11 \pm 0.44$ \\
%     \bottomrule
%     \end{tabular}
%     \label{tab:comp}
% \end{table*}

We evaluate \ourmethod{} in several learning setups using three common image classification datasets:  CIFAR10, CIFAR100, and Omniglot \citep{cifar, lake2015human} %We distribute these datasets across $n$ clients based on a diverse set of learning setups.
Unless stated otherwise, we report the Federated Accuracy, defined as   $\frac{1}{n}\sum_i\frac{1}{m_i}\sum_j \text{Acc}\left(f_i\left(\bx^{(i)}_j\right), y^{(i)}_j\right)$, averaged over three seeds.
The experiments show that \ourmethod{} outperforms classical FL approaches and leading PFL models. 
% We will make our code publicly available to support the reproducibility of our results and future research.

\paragraph{Compared Methods:} We evaluate and compare the following approaches: \textbf{(1)} \textbf{\ourmethod{}}, Our proposed Federated HyperNetworks  \textbf{(2)} \textbf{\ourmethod{}-PC}, \ourmethod{} with a personalized classifier per client ; 
\textbf{(3)} \textbf{Local}, Local training on each client, with no collaboration between clients;
\textbf{(4)} \textbf{FedAvg}~\cite{McMahan2017CommunicationEfficientLO}, one of the first and perhaps the most widely used FL algorithm;
\textbf{(5)} \textbf{Per-FedAvg}~\cite{Fallah2020PersonalizedFL} a meta-learning based PFL algorithm.
 \textbf{(6) pFedMe}~\cite{Dinh2020PersonalizedFL}, a PFL approach which adds a Moreau-envelopes loss term% on the client-side optimization
 ; \textbf{(7) LG-FedAvg}~\cite{liang2020think} PFL method with local feature extractor and global output layers; \textbf{(8) FedPer}~\cite{arivazhagan2019federated} a PFL approach that learns per-client personal classifier on top of a shared feature extractor.

\paragraph{Training Strategies:} In all experiments, our target network shares the same architecture as the baseline models. Our hypernetwork is a simple fully-connected neural network, with three hidden layers and multiple linear heads per target weight tensor. We limit the training process to at-most $5000$ server-client communication steps for most methods. One exception is LG-FedAvg which utilizes a pretrained FedAvg model, hence it is trained with additional $1000$ communication steps. The \textit{Local} baseline is trained for $2000$ optimization steps on each client. For \ourmethod{}, we set the number of local steps to $K=50$, and the embedding dimension to $\lfloor 1+n / 4 \rfloor$, where $n$ is the number of clients. We provide an extensive ablation study on design choices in Appendix~\ref{app:additional_exp}.
We tune the hyperparameters of all methods using a pre-allocated held-out validation set. Full experimental details are provided in Appendix~\ref{app:exp_details}.

\subsection{Heterogeneous Data}\label{sec:hetro}
% We adopt the setup proposed in~\citet{Dinh2020PersonalizedFL} for generating heterogeneous settings in terms of classes and size of local training data, using CIFAR10, CIFAR100 and Omniglot datasets. For CIFAR10/CIFAR100, we 

We evaluate the different approaches on a challenging heterogeneous setup. We adopt the learning setup and the evaluation protocol described  in~\citet{Dinh2020PersonalizedFL} for generating heterogeneous clients in terms of classes and size of local training data.  First, we sample two/ten classes for each client for CIFAR10/CIFAR100; Next, for each client $i$ and selected class $c$, we sample $\alpha_{i,c}\sim U(.4,.6)$, and assign it with $\frac{\alpha_{i,c}}{\sum_j \alpha_{j,c}}$ of the samples for this class. We repeat the above using $10, 50$ and $100$ clients. This procedure produces clients with different number of samples and classes.
%By following this procedure, each client receives a unique number of samples. 
For the target network, we use a LeNet-based~\cite{lecun1998gradient} network with two convolution and two fully connected layers.
%In addition, to previous datasets, we show the superiority of \ourmethod{} on Omniglot dataset. 

We also evaluate all methods using the Omniglot dataset \cite{lake2015human}.
%Omniglot is commonly used for few-shot learning. 
Omniglot contains 1623 different grayscale handwritten characters (with 20 samples each), from 50 different alphabets. Each alphabet obtains a varying number of characters. In this setup, we use 50 clients and assign an alphabet to each client. Therefore, clients receives  different numbers of samples and the distribution of labels is disjoint across clients. We use a LeNet-based model with four convolution and two fully connected layers.

The results are presented in Table~\ref{tab:hetro}. 
%Our approach outperforms all baselines, 
The two simple baselines, local and FedAvg, that do not use personalized federated learning perform quite poorly on most tasks\footnote{In the 10-client split, each client sees on average 10\% of the train set. It is sufficient for training a model locally.}, showing the importance of personalized federated learning. \ourmethod{} achieves large improvements of 2\%-10\% over all competing approaches. Furthermore, on the Omniglot dataset, where each client is allocated with a completely different learning task (different alphabet), we show significant improvement using \ourmethod{}-PC. We present additional results on the MNIST dataset in Appendix~C.%\ref{app:additional_exp}.

\subsection{Computational Budget}\label{sec:comp}

\begin{table*}[t]
    \small
    % \scriptsize
    \centering
    \caption{\textit{Computational budget}. Test accuracy for CIFAR10/100 with 75 clients and varying computational capacities.}
    \vskip 0.11in
    \begin{tabular}{l c c c c c c c c}
    \toprule
    & \multicolumn{3}{c}{CIFAR10} && \multicolumn{3}{c}{CIFAR100} \\
     \cmidrule{2-4} \cmidrule{6-8}\\
    Local model size & S  & M & L && S  & M & L\\
    \midrule
    FedAvg & $36.91 \pm 2.26$ & $42.09 \pm 2.37$ & $47.51\pm 1.97$ && $9.79 \pm 0.19$ & $11.76 \pm 1.14$ & $17.86 \pm 2.42$\\
    Per-FedAvg & $70.72 \pm 2.57$ & $72.33 \pm 2.57$ & $75.13 \pm 0.61$ && $41.49 \pm 0.91$ & $43.22 \pm 0.16$ & $44.03 \pm 1.77$ \\
    pFedMe & $81.21 \pm 1.23$ & $84.08 \pm 1.63$ & $83.15 \pm 2.45$   && $39.91 \pm 0.81$ & $41.99 \pm 0.55$ & $44.93 \pm 1.63$ \\
    LG-FedAvg & $73.93 \pm 3.65$ & $53.13 \pm 3.49$ & $54.72 \pm 2.50$ && $33.48 \pm 4.83$ & $29.15 \pm 1.51$ & $23.01 \pm 1.41$\\
    \midrule
    \ourmethod{} (ours) & $\mathbf{85.38 \pm 1.21}$ & $\mathbf{86.92 \pm 1.35}$ & $\mathbf{87.20 \pm 0.76}$ && $\mathbf{48.04 \pm 0.89}$ & $\mathbf{48.66 \pm 1.21}$ & $\mathbf{50.66 \pm 2.70}$\\
    %\midrule
    % \ourmethod{} (ours S) & $87.80 \pm 0.1$ & $-$ & $-$ && $52.24 \pm 0.71$ & $-$ & $-$ \\
    % \ourmethod{} (ours M) & $-$ & $ 88.83 \pm 0.14 $ & $-$ && $-$ & $55.48 \pm 0.90$ & $-$ \\
    %\ourmethod{} (ours L) & $-$ & $-$ & $89.06 \pm 0.14$ && $-$ & $-$ & $56.11 \pm 0.44$ \\
    \bottomrule
    \end{tabular}
    \label{tab:comp}
\end{table*}

% \begin{table*}[t]
%     \vskip 0.15in
%     % \small
%     \scriptsize
%     \centering
%     \caption{Computational budget.}
%     \begin{tabular}{l c c c c c c c c}
%     \toprule
%     & \multicolumn{3}{c}{CIFAR10} && \multicolumn{3}{c}{CIFAR100} \\
%      \cmidrule{2-4} \cmidrule{6-8}\\
%      & S  & M & L && S  & M & L\\
%     \midrule
%     FedAvg & $36.91 \pm 2.26$ & $42.09 \pm 2.37$ & $47.51\pm 1.97$ \\
%     Per-FedAvg & $70.72 \pm 2.57$ & $72.33 \pm 2.57$ & $75.13 \pm 0.61$ && $41.49 \pm 0.91$ & $43.22 \pm 0.16$ & $44.03 \pm 1.77$ \\
%     pFedMe & $81.21 \pm 1.23$ & $84.08 \pm 1.63$ & $83.15 \pm 2.45$   && $39.91 \pm 0.81$ & $41.99 \pm 0.55$ & $44.93 \pm 1.63$ \\
%     LG-FedAvg & $73.93 \pm 3.65$ & $53.13 \pm 3.49$ & $54.72 \pm 2.50$ \\
%     \midrule
%     \ourmethod{} (ours) & $\mathbf{85.38 \pm 1.21}$ & $\mathbf{86.92 \pm 1.35}$ & $\mathbf{87.20 \pm 0.76}$ \\
%     \midrule
%     \ourmethod{} (ours S) & $87.80 \pm 0.1$ & $-$ & $-$ \\
%     \ourmethod{} (ours M) & $-$ & $ 88.83 \pm 0.14 $ & $-$ \\
%     \ourmethod{} (ours L) & $-$ & $-$ & $89.06 \pm 0.14$ \\
%     \bottomrule
%     \end{tabular}
%     \label{tab:comp}
% \end{table*}

We discussed above how the challenges of heterogeneous data can be handled using \ourmethod{}. Another major challenge presented by personalized FL is that the communication, storage, and computational resources of clients may differ significantly. These capacities may even change in time due to varying network and power conditions.
In such a setup, the server should adjust to the communication and computational policies of each client. Unfortunately, previous works do not address this resource heterogeneity. \ourmethod{} can naturally adapt to this challenging learning setup by producing target networks of different sizes. 

In this section, we evaluate the capacity of \ourmethod{} to handle clients that differ in their computational and communication resource budget. We use the same split described in Section~\ref{sec:hetro} with a total of $75$ clients divided into the three equal-sized groups named \textit{S} (small), \textit{M} (medium), and \textit{L} (large). The Models of clients within each group share the same architecture. The three architectures (of the three groups) have a different number of parameters. 

We train a single \ourmethod{} to output target client models of different sizes. Importantly, this allows \ourmethod{} to share parameters between all clients even if those have different local model sizes. 

\noindent\textbf{Baselines:} For quantitative comparisons, and since the existing baseline methods cannot easily extend to this setup, we train three independent per-group models. Each group is trained for $5000$ server-client communication steps. 
% We also compare \ourmethod{} to an upper-bound baseline based on our method, denoted \ourmethod{}(L), which generates three large-sized models. 
See details in Appendix~C %\ref{app:exp_details}
for further details.

The results are presented in Table~\ref{tab:comp}, showing that \ourmethod{} achieves $4\%-8\%$ improvement over all competing methods. The results demonstrate the flexibility of our approach, which is capable of adjusting to different client settings while maintaining high accuracy.

\subsection{Generalization to Novel Clients}

% \begin{figure}[ht]
%     \centering
%     \includegraphics[width=1.\linewidth]{figures/generalization_tv.png}
%     \caption{\textit{Generalization to novel clients}. The generalization gap accuracy between training and novel clients, defined as $\text{acc}_{novel} - \text{acc}_{train}$, where $\text{acc}$ denotes the average accuracy.}
%     \label{fig:gen}
% \end{figure}

Next, we study an important learning setup where
new clients join, and a new model has to be trained for their data. In the general case of sharing models across clients, this would require retraining (or finetuning) the shared model. While PFL methods like pFedME~\cite{Dinh2020PersonalizedFL} and Per-FedAvg~\cite{Fallah2020PersonalizedFL} can adapt to this setting by finetuning the global model locally, \ourmethod{} architecture offers a significant benefit. Since the shared model learns a meta-model over the distribution of clients, it can in principle generalize to new clients without retraining. 
%The PFL environment is ever-changing and dynamic, with new clients joining the network. These challenging environment conditions must be accounted for when designing a PFL algorithm. Here we evaluate the generalization performance of the different PFL methods when applied to novel clients. More concretely, consider that a PFL model was trained on a set of training clients to convergence. 
%We now wish to evaluate the performance of clients outside of the training population while fixing the global model (shared parameters).
With \ourmethod{}, once the shared model $\varphi$ has been trained on a set of clients, extending to a new set of novel clients requires little effort. We freeze the hypernetwork weights $\varphi$ and optimize an embedding vector $\bv_{new}$.
%, and personalized classifier for each of the novel clients, using their personal data. 
Since only a small number of parameters are being optimized, training is less prone to overfitting compared to other approaches. The success of this process depends on the capacity of the hypernetwork to learn the distribution over clients and generalize to clients that have different data distributions. 

\begin{figure}[ht]
    \centering
    \includegraphics[width=1.\linewidth]{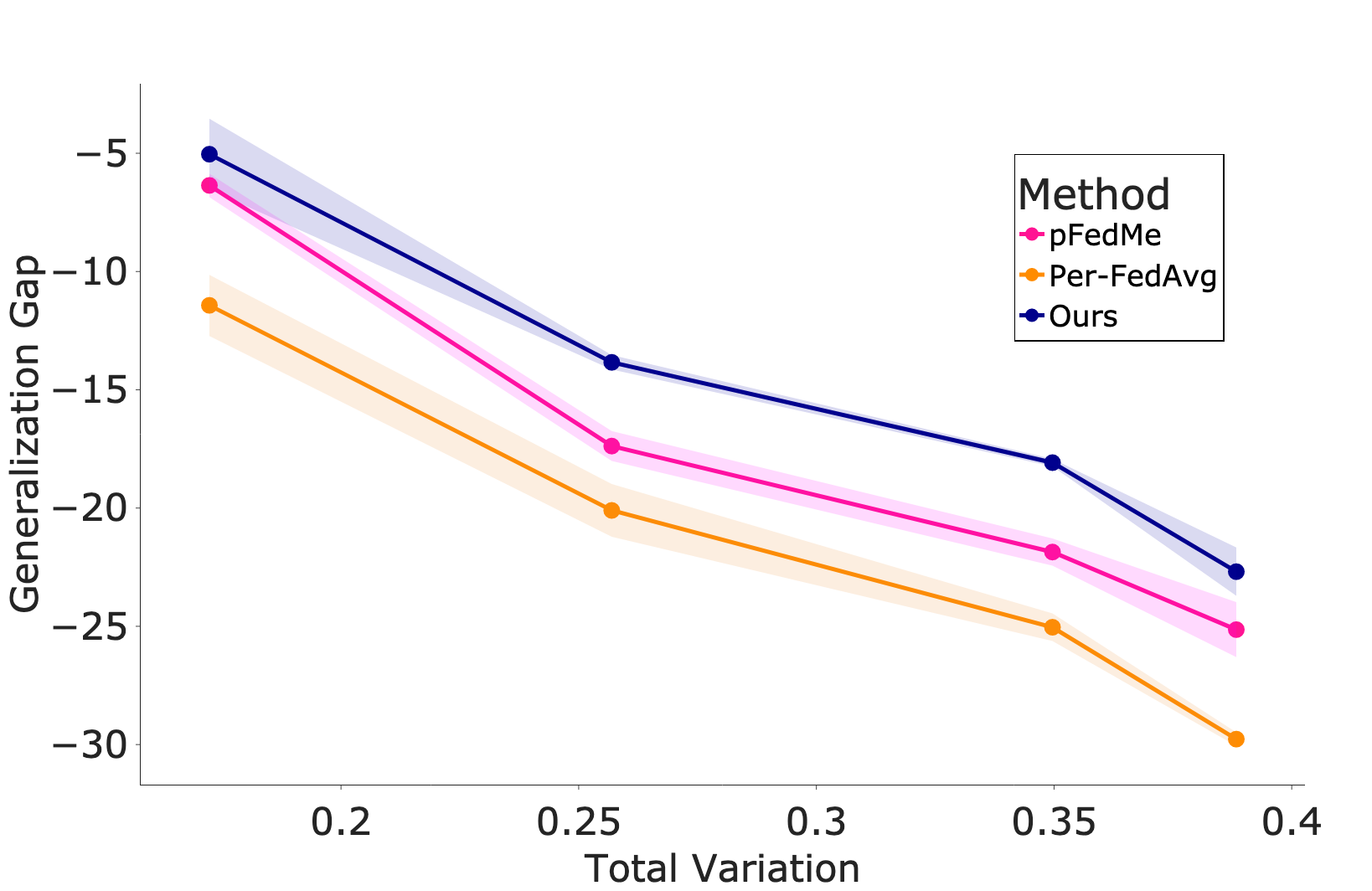}
    \caption{\textit{Generalization to novel clients}. The accuracy generalization gap between training and novel clients, defined as $\text{acc}_{novel} - \text{acc}_{train}$, where $\text{acc}$ denotes the average accuracy.}
    \label{fig:gen}
\end{figure}

%We first describe how \ourmethod{} naturally allows us to incorporate novel clients. Given a novel client, we can freeze the hypernetwork weights and optimize the novel clients embedding vector $\bv$ using its data.
%Assuming $\bv$ is informative enough, we should expect the corresponding personal model to perform well with no additional fine-tuning.  However, if $\bv$ is an embedding vector, we can fine-tune it on the novel client's data. When fitting the embedding vector of  novel clients we first initialize it to the average over the embedding vectors of the training clients. Then we fine-tune it by using the novel client's data while freezing the HN weights. As we only optimizing a small number of parameters, we are less prone to overfitting compared to other approaches.
To evaluate \ourmethod{} in this setting, we use the CIFAR10 dataset, with a total of 100 clients, of which 90 are used for training and 10 are held out novel clients. To allocate data samples, for each client $i$ we first draw a sample from a Dirichlet distribution with parameter $\boldsymbol{\alpha}=(\alpha, ...,\alpha)\in \mathbb{R}^{10}$, $p_i\sim Dir(\boldsymbol{\alpha})$. Next we normalize the $p_i$'s so that $\sum_i p_{i,j}=1$ for all $j$, to obtain the vector $\hat{p}_i$. We now allocate samples according to the $\hat{p}_i$'s. For the training clients, we choose $\alpha=.1$, whereas for the novel clients we vary $\alpha\in\{.1, .25, .5, 1\}$. To estimate the ``distance'' between a novel client and the training clients, we use the total variation (TV) distance between the novel client and its nearest neighbor in the training set. The TV is computed over the empirical distributions $\hat{p}$. Figure~\ref{fig:gen} presents the accuracy generalization gap as a function of the total variation distance. \ourmethod{} achieves the best generalization performance for all levels of TV (corresponds to the different values for $\mathbb{\alpha}$).

%We tested this idea using CIFAR10 dataset 
%with the same data split described in Section~\ref{sec:hetro} 
%with 100 clients, where 90 clients are used for training, and 10 novel clients. 

% We train \ourmethod{} with local layers and learn the embedding vectors and the local layers novel clients. For the other PFL methods, we follow the training and evaluation protocol used for evaluating local models on training clients, performing the same number of local optimization steps.~\an{we need better explanation/phrasing here}. \as{need to write the learning setup including data splits}

\subsection{Heterogeneity of personalized classifiers}

\begin{figure}[ht]
    \centering
    \includegraphics[width=0.9\linewidth]{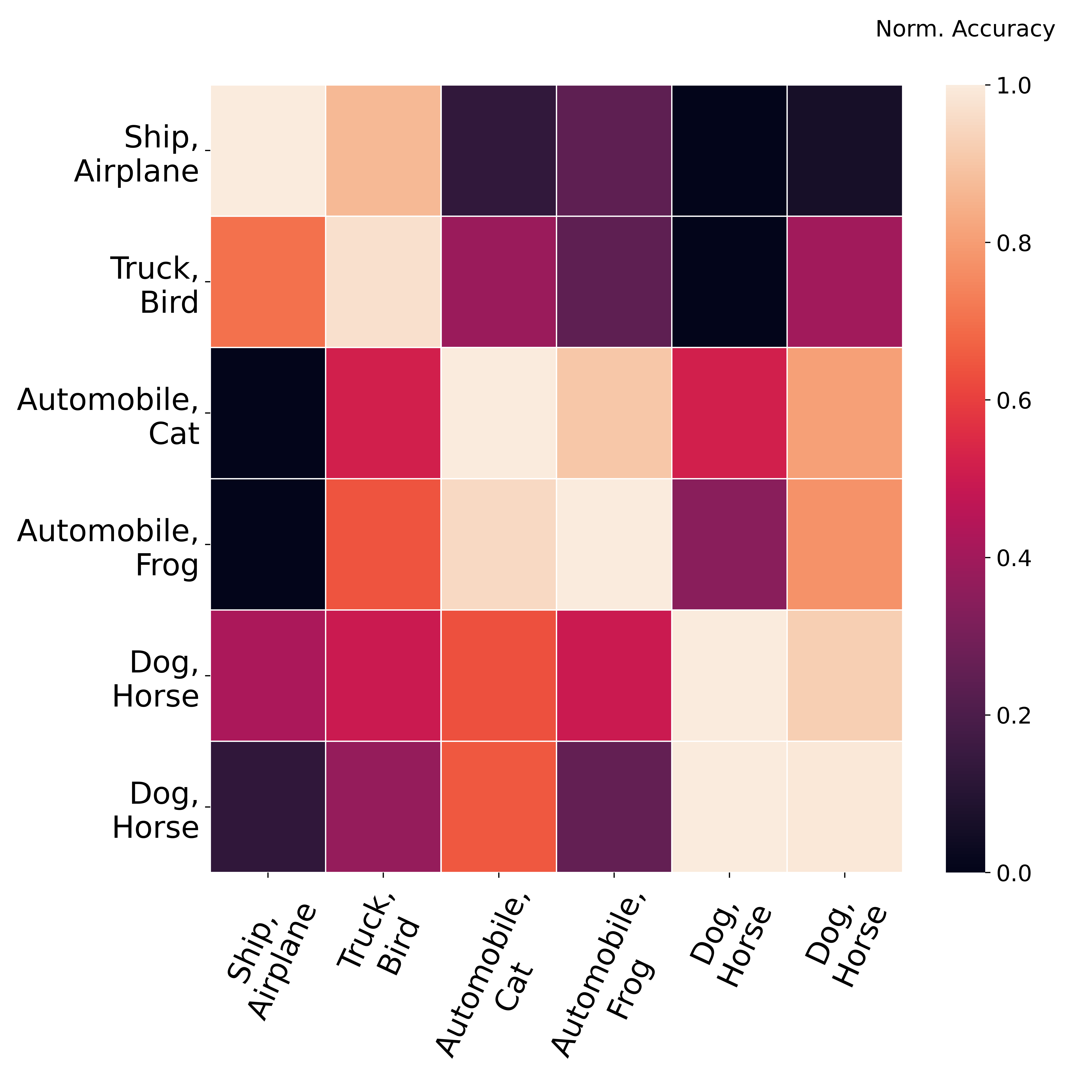}
    \caption{\textit{Model personalization}. Rows correspond to clients, each with their trained in a binary classification task and keeping their personalized classifier $\omega_i$. Columns correspond to the feature extractor $\theta_j$ of another client. The diagonal corresponds to the stanard training with $\theta_i$ and $\omega_i$. For better visualization, values denote accuracy normalized per row:  $\text{norm-acc}_{i,j}=(\text{acc}_{i,j}-\min_{\ell}{\text{acc}_{i,\ell}}) / (\max_{\ell}{\text{acc}_{i,\ell}}-\min_{\ell}{\text{acc}_{i,\ell}})$.}
    \label{fig:personalized_fe}
\end{figure}

% We further investigate the flexibility of \ourmethod{}-PC in terms of personalizing different networks for different clients. We wanted to investigate how different where the personalized feature extractors we generated to exclude the chance that only the classification layer was personalized. Intuitively, this is not the case since if it was, we would have seen similar performance to FedPer~\cite{arivazhagan2019federated} as the clients would share a global feature extractor. We then investigate: do the feature extraction layers generated by the \ourmethod{}-PC's hypernetwork actually differ from each other?

We further investigate the flexibility of \ourmethod{}-PC in terms of personalizing different networks for different clients. 
Potentially, since clients have their own personalized classifier $\omega_i$, the feature extractor component $\theta_i$ generated by the HN may in principle become strongly similar across clients, making the HN redundant. %shared the learned personal models might share a feature extractor and differ only in the classification layers. 
The empirical results show that this is not the case because \ourmethod{}-PC out-performs FedPer~\cite{arivazhagan2019federated}. However this raises a fundamental question about the interplay between local and shared personalized components. We provide additional insight to this topic by answering the question: do the feature extraction layers generated by the \ourmethod{}-PC's hypernetwork significantly differ from each other?

%yPotentially, the learned personal models might share a feature extractor and differ only in the classification layers. Intuitively, this is not the case since if it were, we would have seen similar performance to FedPer~\cite{arivazhagan2019federated}. We provide additional insight by answering the question: do the feature extraction layers generated by the \ourmethod{}-PC's hypernetwork actually differ from each other?

To investigate the level of personalization in $\theta_i$ achieved by \ourmethod{}-PC, we first train it on CIFAR10 dataset split among ten clients, with two classes assigned to each client. Next, for each client, we replace its feature extractor $\theta_i$ with that of another client $\theta_j$ while keeping its personal classifier $\omega_i$ unaltered. 
Figure~\ref{fig:personalized_fe} depicts the normalized accuracy in this mix-and-match experiment. Rows correspond to a client, and columns correspond to the feature extractor of another client. 

Several effects in Figure~\ref{fig:personalized_fe} are of interest. First, \ourmethod{}-PC produces personalized feature extractors for each client since the accuracy achieved when crossing classifiers and feature extractors varies significantly. Second, some pairs of clients can be crossed without hurting the accuracy. Specifically, we had two clients learning to discriminate \textit{horse} vs. \textit{dog}. Interestingly, the client with \textit{ship} and \textit{airplane} classes performs quite well when presented with \textit{truck} and \textit{bird}, presumably because both of their feature extractors learned to detect sky.

\subsection{Learned Client Representation}

\begin{figure}[ht]
    \centering
    \includegraphics[width=1.\linewidth]{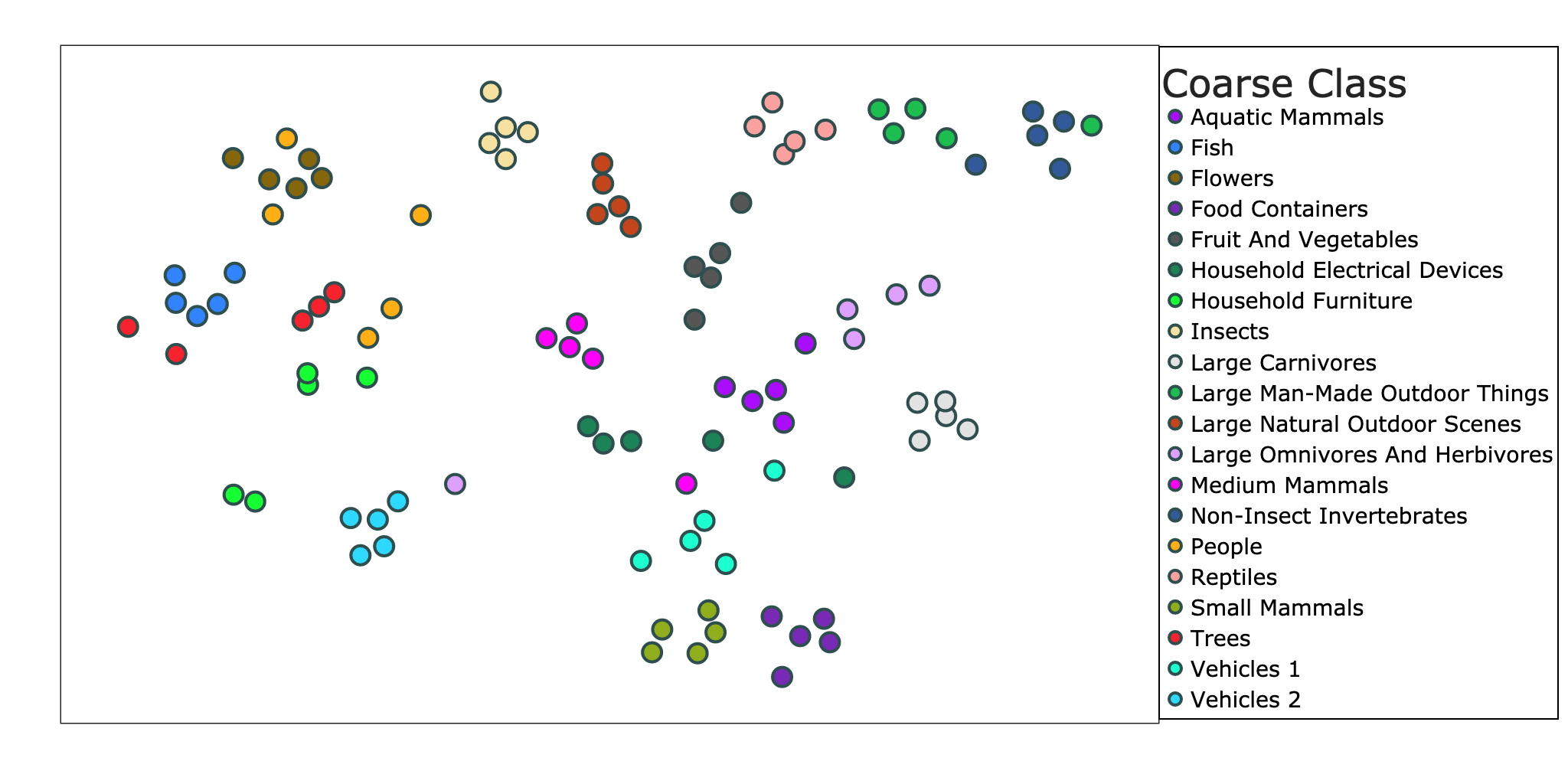}
    \caption{t-SNE visualization of the learned client representation $\bv$ for the CIFAR100 dataset. Clients are tasked with classifying classes that belong to the same coarse class. Clients marked with the same color correspond to the same coarse-class, see text for details. \ourmethod{} clustered together clients from the same group.}
    \label{fig:embedding}
\end{figure}
% SORRY, GOOD IDEA TO EDIT HERE, GO AHEAD :) 
%In some cases, a meaningful client representation is available and can be used as an input for the HN. Otherwise, we initialize a random embedding vector for each client, and update it throughout the optimization process similar to the HN parameters (see Alg.~\ref{alg:fhn}). 
% Here we try to see if the embedding learned per client learns a meaningful representation of that client. 
In our experiments, we learn to represent each client using a trainable embedding vector $v_i$. These embedding vectors therefore learn a continuous semantic representation over the set of clients. The smooth nature of this representation gives the HN the power to share information across clients. 
We now wish to study the structure of that embedding space.  

To examine how the learned embedding vectors reflect a meaningful representation over the client space, 
%Can we learn meaningful representation in the latter case? 
we utilize the hierarchy in CIFAR100 for generating clients with similar data distribution of semantically similar labels. Concretely, we split the CIFAR100 into 100 clients, where each client is assigned with data from one out of the twenty coarse classes uniformly (i.e., each coarse class is assigned to five clients). 

In Figure~\ref{fig:embedding} we project the learned embedding vectors into $\mathbb{R}^2$ using the t-SNE algorithm~\cite{Maaten2008VisualizingDU}. A clear structure is presented, in which clients from the same group (in terms of coarse labels) are clustered together. 

\section{Conclusion}

In this work, we present a novel approach for personalized federated learning. Our method trains a central hypernetwork to output a unique personal model for each client.
We show through extensive experiments significant improvement in accuracy on all datasets and learning setups.

Sharing across clients through a central hypernetwork has several benefits compared to previous architectures. First, since it learns a unified model over the distribution of clients, the model generalizes better to novel clients, without the need to retrain the central model. Second, it naturally extends to handle clients with different compute power, by generating client models of different sizes. 
Finally, this architecture decouples the problem of training complexity from communication complexity, since local models that are transmitted to clients can be significantly more compact than the central model.

We expect that the current framework can be further extended in several important ways. First, the architecture opens questions about the best way of allocating learning capacity to a central model vs distributed components that are trained locally. Second, the question of generalization to clients with new distribution awaits further analysis.

% \iffalse
% \subsection{Using meta data}

% Another motivation for using \ourmethod{} is the ability to better utilize metadata without compromising privacy. In some systems, we have prior knowledge of certain groups of clients. In healthcare, we might have metadata on the measuring device, such as its manufacturer or its power consumption. Such metadata might lead to better performance since they provide implicit information about the data used by a specific user group. \as{cite modularity of HN} supports this motivation as they showed the superiority of HN’s modularity i.e. the ability to effectively learn a different function for each input instance over embedding-based methods. In this experiment we use the hierarchy of Omniglot as the metadata. Each client is assigned labels from ?? languages. .... \as{further describe the learning setup and which method we used for pFedMe. In which stage of the forward pass "fed" the network with the metadata}

% \an{Hopefully we can add more experiments here :)}

% \fi

\nocite{langley00}

\section*{Acknowledgements}
This study was funded by a grant to GC from the Israel Science Foundation (ISF 737/2018), and by an equipment grant to GC and Bar-Ilan University from the Israel Science Foundation (ISF 2332/18). AS and AN were funded by a grant from the Israeli Innovation Authority, through the AVATAR consortium.

\bibliography{ref}
\bibliographystyle{icml2021}

\clearpage
\twocolumn[
\icmltitle{Supplementary Material for Personalized Federated Learning by Hypernetworks}]

\appendix

% \twocolumn[{%
% \renewcommand\twocolumn[1][]{#1}%
% % \maketitle
% \begin{center}
%     \LARGE{Personalized Federated Learning by Hypernetworks}
% \end{center}%
% }]
% \begin{center}
% {\LARGE{Personalized Federated Learning by Hypernetworks}}
% \end{center}

\section{Proof of Results}\label{app:theory}

\paragraph{Proof for Proposition 1.} Let $\bar{\theta}_i$ denote the optimal solution at client $i$, then $\bar{\theta}_i=(X_i^TX_i)^{-1}X_i^T\by_i=X_i^T\by_i$. Denote $\theta_i=W\bv_i$, we have 
\begin{align*}
    &\arg\min_{\theta_i}\Vert X_i\theta_i-\by_i\Vert_2^2 =\arg\min_{\theta_i}(X_i\theta_i-\by_i)^T(X_i\theta_i-\by_i)\\
    &=\arg\min_{\theta_i}\theta_i^TX_i^TX_i\theta_i-2\theta_i^TX_i^T\by+\by_i^T\by_i\\    
    &=\arg\min_{\theta_i}\theta_i^T\theta_i-2\langle\theta_i,\bar{\theta}_i\rangle+\by_i^T\by_i\\
    &=\arg\min_{\theta_i} \theta_i^T\theta_i-2\langle\theta_i,\bar{\theta}_i\rangle+\Vert\bar{\theta}_i\Vert_2^2\\%-\Vert\bar{\theta}_i\Vert_2^2\\
    &=\arg\min_{\theta_i} \Vert\theta_i-\bar{\theta}_i\Vert_2^2
\end{align*}
Thus, our optimization problem becomes $\arg\min_{W,V}\sum_i\Vert W\bv_i-\bar{\theta}_i\Vert_2^2$. 

WLOG, we can optimize $W$ over the set of all matrices with orthonormal columns, i.e. $W^TW=I$. Since for each solution $(W,V)$ we can obtain the same loss for $(WR,R^{-1}V)$, and select a $R$ that performs Gram-Schmidt on the columns of $W$. In case of fixed $W$ the optimal solution for $
\bv_i$ is given by $\bv_i^*=(W^TW)^{-1}W^T\bar{\theta}_i=W^T\bar{\theta}_i$. Hence, our optimization problem becomes,
\begin{align*}
    &\arg\min_{W; W^TW=I} \sum_i \Vert WW^T\bar{\theta}_i - \bar{\theta}_i\Vert_2^2,\\
\end{align*}
which is equivalent to PCA on $\{\bar{\theta}_i\}_i$.
\qed

\paragraph{Proof for Theorem 1.}
We note a $\log(1/\epsilon)$ factor missing in the statement of Theorem 1 in the paper, the correct statement should use $M=\mathcal{O}\left(\frac{k}{\epsilon^2}\log\left(\frac{RL(L_h+L_V)}{\epsilon\delta}\right)+\frac{N}{n\epsilon^2}\log\left(\frac{RL(L_h+L_V)}{\epsilon\delta}\right)\right)$.

Using Theorem 4 from \cite{baxter2000model} and the notation used in that paper, we get that $M=\mathcal{O}\left(\frac{1}{n\epsilon^2}\log\left(\frac{\mathcal{C}(\epsilon,\mathds{H}^n_l)}{\delta}\right)\right)$ where $\mathcal{C}(\epsilon,\mathds{H}^n_l)$ is the covering number for $\mathds{H}^n_l$. In our case each element of $\mathds{H}^n_l$ is parametrized by $\varphi,\bv_1,...,\bv_n$ and the distance is given by 
{\small
\begin{align}
    &d((\varphi,\bv_1,...,\bv_n),(\varphi',\bv'_1,...,\bv'_n))=\\
    &\underset{x_i,y_i\sim P_i}{\mathds{E}}\left[\frac{1}{n}\left|\sum \ell(h(\varphi,\bv_i)(x_i),y_i)-\sum \ell(h(\varphi',\bv'_i)(x_i),y_i)\right|\right]\nonumber
\end{align}
}%

From the triangle inequality and our Lipshitz assumptions we get
\begin{align}
    &d((\varphi,\bv_1,...,\bv_n),(\varphi',\bv'_1,...,\bv'_n))\leq \\
    & \sum\frac{1}{n}\underset{x_i,y_i
    \sim P_i}{\mathds{E}}\left[\left| \ell(h(\varphi,\bv_i)(x_i),y_i)- \ell(h(\varphi',\bv'_i)(x_i),y_i)\right|\right] \nonumber \\
    & \leq L\Vert h(\varphi,\bv_i)-h(\varphi',\bv'_i)\Vert \nonumber \\
    &\leq L\Vert h(\varphi,\bv_i)-h(\varphi,\bv'_i)\Vert+ L\Vert h(\varphi,\bv'_i)-h(\varphi',\bv'_i)\Vert \nonumber \\
    &\leq L\cdot L_h\Vert\varphi-\varphi'\Vert+L\cdot L_V\Vert\bv-\bv'\Vert \nonumber
\end{align}
Now if we select a covering of the parameter space such that each $\varphi$ has a point $\varphi'$ that is $\frac{\epsilon}{2L(L_h+L_V)}$ away and each embedding $\bv_i$ has an embedding $\bv_i'$ at the same distance we get an $\epsilon$-covering in the $d((\varphi,\bv_1,...,\bv_n),(\varphi',\bv'_1,...,\bv'_n))$ metric. From here we see that $\log(\mathcal{C}(\epsilon,\mathds{H}^n_l))=\mathcal{O}\left((n\cdot k+N)\log \left(\frac{RL(L_V+L_h)}{\epsilon}\right)\right)$.\qed

\section{Experimental Details}\label{app:exp_details}

For all experiments presented in the main text, we use a fully-connected hypernetwork with $3$ hidden layers of $100$ hidden units each. For all relevant baselines, we aggregate over $5$ clients at each round. We set $K=3$ ,i.e., $60$ local steps, for the pFedMe algorithm, as it was reported to work well in the original paper~\citep{Dinh2020PersonalizedFL}.

\paragraph{Heterogeneous Data (Section 5.1).} 
%Here we provide  full experimental details for the \textit{Heterogeneous Data} experiment presented in Section 5.1
%~\ref{sec:hetro}. 
For the CIFAR experiments, we pre-allocate $10,000$ training examples for validation. For the Omniglot dataset, we use a 70\%/15\%/15\% split for train/validation/test sets. 
The validation sets are used for hyperparameter tuning and early stopping. We search over learning-rate $\{1e-1, 5e-2, 1e-2, 5e-3\}$, and personal learning-rate $\{5e-2, 1e-2, 5e-3, 1e-3\}$ for PFL methods using $50$ clients. For the CIFAR datasets, the selected hyperparameters are used across all number of clients (i.e. $10, 50, 100$). 

\paragraph{Computational Budget (Section 5.2)} We use the same hyperparameters selected in Section 5.1. To align with previous works \cite{Dinh2020PersonalizedFL, liang2020think, Fallah2020PersonalizedFL}, we use a LeNet-based (target) network with two convolution layers, where the second layer has twice the number of filters in comparison to the first. Following these layers are two fully connected layers that output logits vector. In this learning setup, we use three different sized target networks with different numbers of filters for the first convolution layer. Specifically, for $S/M/L$ sized networks, the first convolution layer consists of $8/16/32$ filters, respectively. \ourmethod{}'s HN produces weights vector with size equal to the sum of the weights of the three sized networks combined. Then it sends the relevant weights according to the target network size of the client.

% \paragraph{Generalization to Novel Clients.} 

\section{Additional Experiments}\label{app:additional_exp}

\subsection{MNIST} \label{app:mnist}

We provide additional experiment over MNIST dataset. We follow the same data partition procedure as in the CIFAR10/CIFAR100 heterogeneity experiment, described in Section 5.1.% \ref{sec:hetro}.

For this experiment we use a single hidden layer fully-connected (FC) hypernetwork. The main network (or target network in the case of \ourmethod{}) is a single hidden layer FC NN.

All FL/PFL methods achieve high classification accuracy on this dataset, which makes it difficult to attain meaningful comparisons. The results are presented in Table~\ref{tab:mnist_hetro}. \ourmethod{} achieves similar results to pFedMe.

\begin{table}[th]
    \vskip 0.15in
    % \small
    \tiny
    \centering
    \caption{Comparison on the MNIST dataset.}
    \begin{tabular}{l c c c }
    \toprule
    & \multicolumn{3}{c}{MNIST}\\
     \cmidrule{2-4}\\
     & 10  & 50 & 100\\
    \midrule
    FedAvg & $96.22\pm 0.65$ & $97.12\pm 0.07$ & $96.99\pm 0.19$\\
    Per-FedAvg & $97.72\pm 0.05$ & $98.57\pm 0.07$ & $98.75\pm 0.26$\\
    pFedMe & $99.40 \pm 0.04$ & $99.30 \pm 0.13$ & $99.12\pm 0.06$\\
    \midrule
    \ourmethod{} (ours) & $99.53 \pm 0.16$ & $99.28 \pm 0.11$ & $99.16 \pm 0.19$\\
    \bottomrule
    \end{tabular}
    \label{tab:mnist_hetro}
\end{table}

\subsection{Exploring Design Choices}
% \ef{Not sure about the name, but I don't have any better suggestions}

In this section we return to the experimental setup of Section~5.1, 
%\ref{sec:hetro}
and evaluate \ourmethod{} using CIFAR10 dataset with 50 clients. First, we examine the effect of the local optimization steps. Next, we vary the capacity of the HN and observe the change in classification accuracy. Finally we vary the dimension of the client representation (embedding).

\subsubsection{Effect of Local Optimization}

\begin{figure}[h]
    \centering
    \includegraphics[width=1.\linewidth]{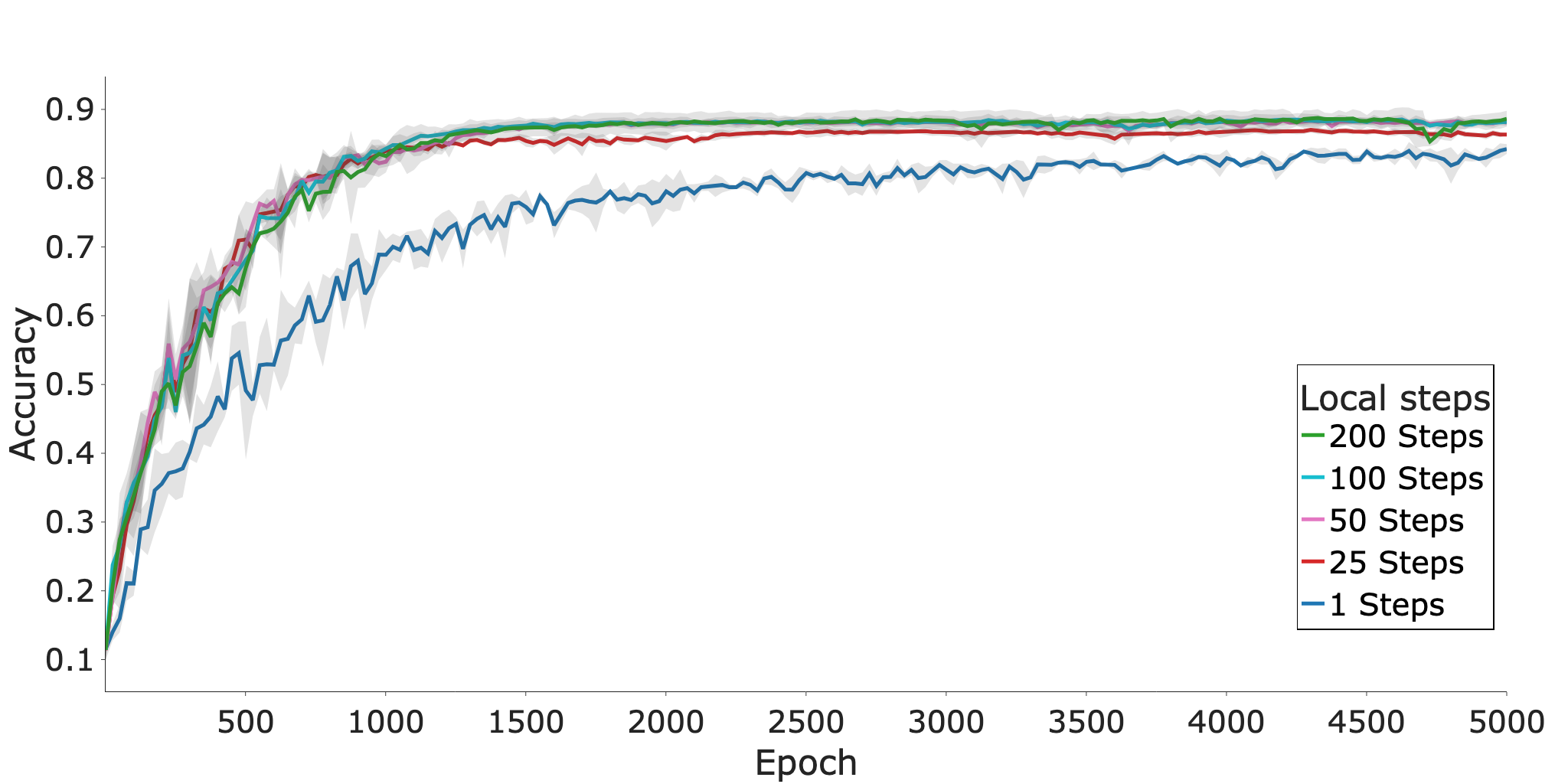}
    \caption{Effect of the number of local optimization steps on the test accuracy for the CIFAR10 dataset.}
    \label{fig:local_opt}
\end{figure}

First, we examine the effect of performing local optimization step and transmitting $\Delta\theta$ back to the hypernetwork. Figure~\ref{fig:local_opt} shows the test accuracy throughout the training process. 
It compares training using the standard chain rule ($\text{steps}=1$) with the case of training locally for $k$ steps, $k\in\{25,50,100,200\}$.
Using our proposed update rule, i.e., making multiple local update steps, yields large improvements in both convergence speed and final accuracy, compared to using the standard chain rule (i.e., $k=1$). The results show that \ourmethod{} is relatively robust to the choice of local local optimization steps. As stated in the main text we set $k=50$ for all experiments.

\subsubsection{Client Embedding Dimension}
\begin{figure*}[t]
\centering
    \begin{subfigure}[]{
    \includegraphics[width=0.45\linewidth]{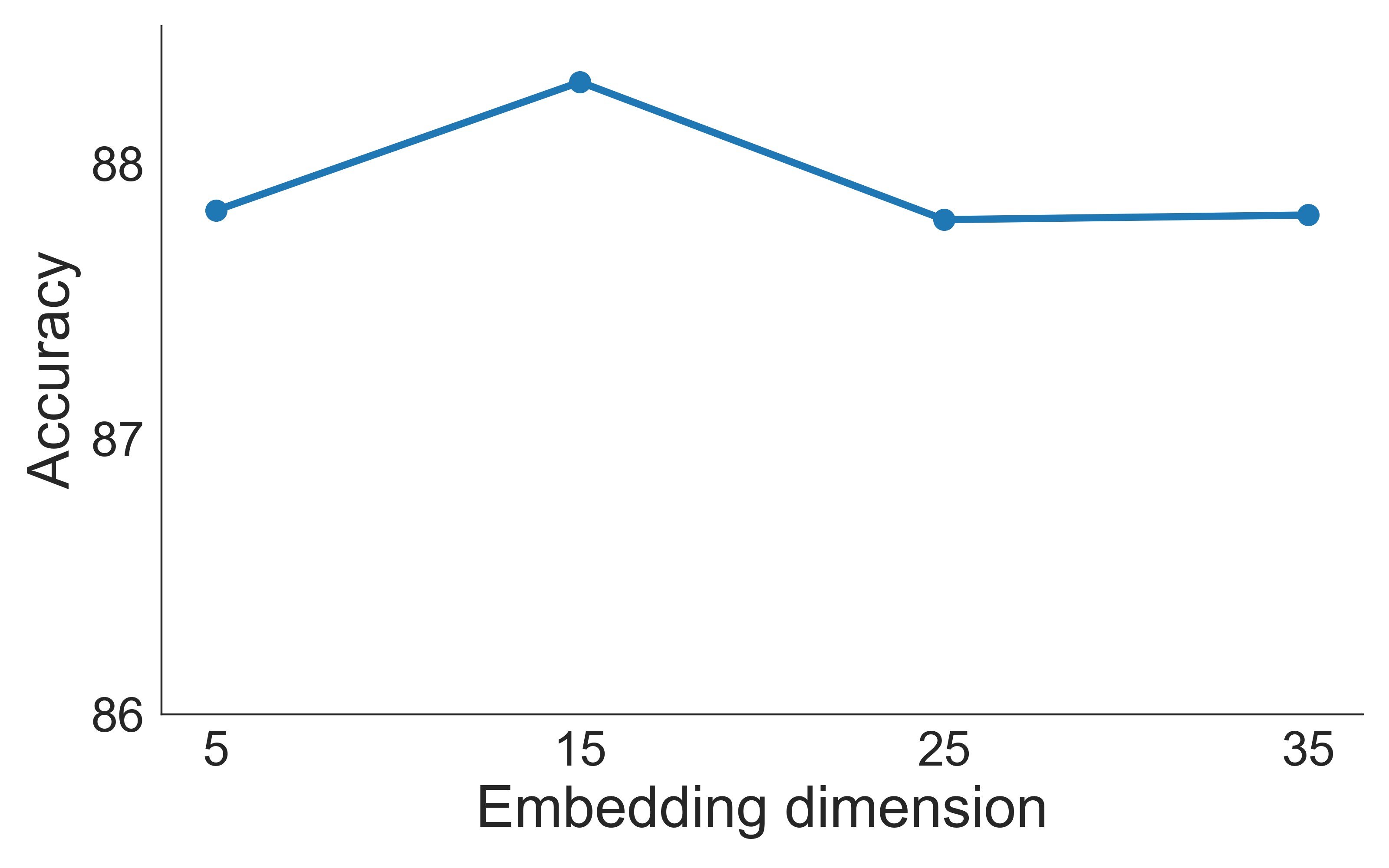}
    }
    \end{subfigure}
    \hfill
    \begin{subfigure}[]{
    \includegraphics[width=0.45\linewidth]{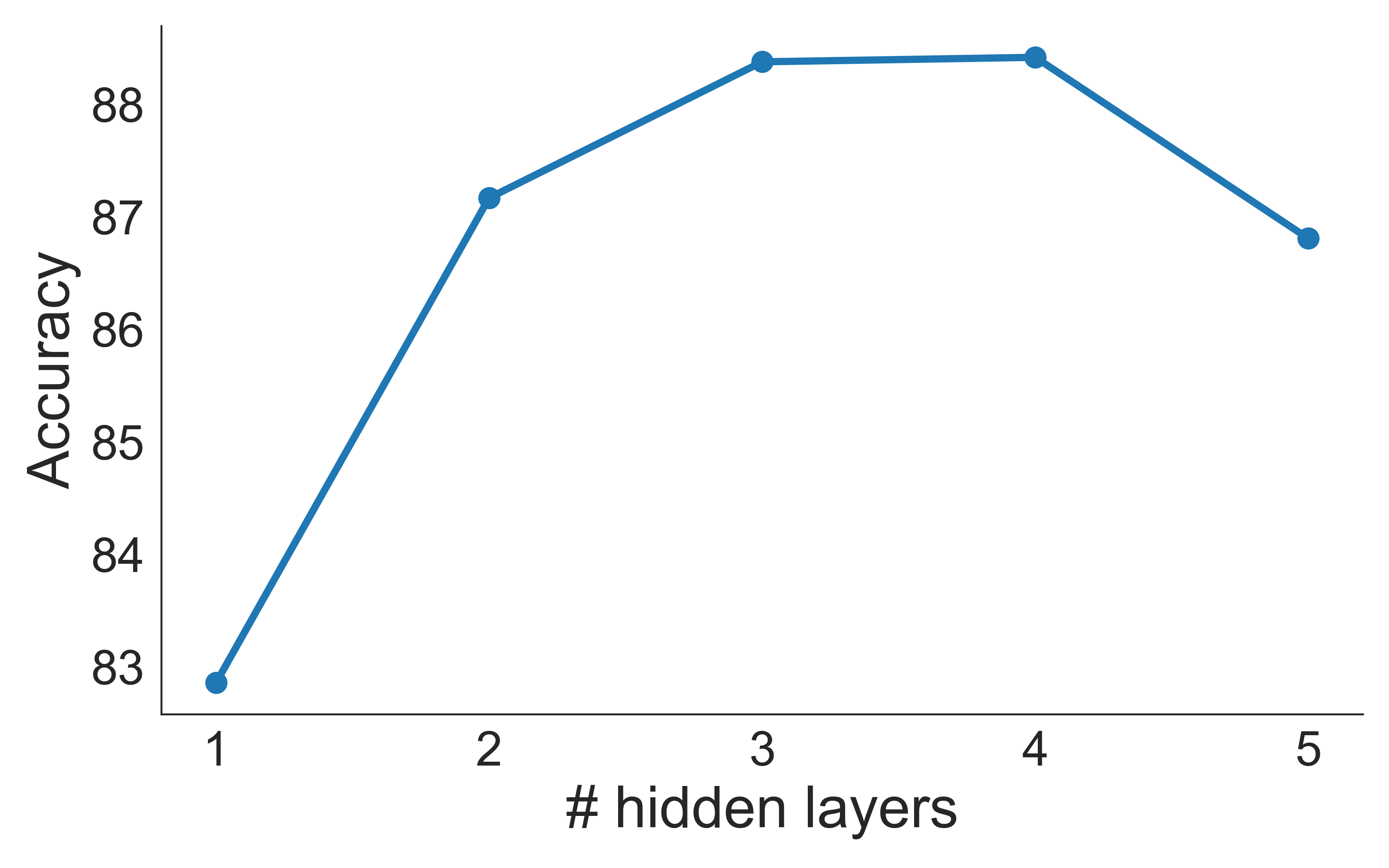}
    }
     \end{subfigure}
    \caption{Test results on CIFAR10 showing the effect of (a) the dimension of the the client embedding vector, and; (b) the number of hypernetwork's hidden layers.}
    \label{fig:ablation}
\end{figure*}

Next, we investigate the effect of embedding vector dimension on \ourmethod{} performance. Specifically, we run an ablation study on set of different embedding dimensions $\{5,15,25,35\}$. The results are presented in Figure~\ref{fig:ablation} (a). We show \ourmethod{} robustness to the dimension of the client embedding vector; hence we fix the embedding dimension through all experiments to $\lfloor 1+n / 4 \rfloor$, where $n$ is the number of client.

% Link to W\&B for results: \url{https://wandb.ai/ax2/fhn/sweeps/ynemwf7z}

\subsubsection{Hypernetwork Capacity}
Here we inspect the effect of the HN's capacity on the local networks performance. We conducted an experiment in which we change the depth of the HN by stacking fully connected layers.

We evaluate \ourmethod{} on CIFAR10 dataset using $k\in\{1,2,3,4,5\}$ hidden layers. Figure~\ref{fig:ablation} (b) presents the final test accuracy. \ourmethod{} achieves optimal performance with $k=3$ and $k=4$ hidden layers, with accuracies $88.38$ and $88.42$ respectively. We use a three hidden layers HN for all experiments in the main text.
% Link to W\&B for results: \url{https://wandb.ai/ax2/fhn/sweeps/h0rsz4im}

\subsection{Spectral Normalization}\label{app:specnorm}

\renewcommand{\tabcolsep}{3pt}
\begin{table}[h]
    \vskip 0.15in
    \small
    %\tiny
    \centering
    %\scalebox{0.9}{
    \caption{\ourmethod{} with spectral-normalization.}
    \begin{tabular}{l c c c }
    \toprule
    & \multicolumn{3}{c}{CIFAR10}\\
     \cmidrule{2-4}\\
     & 10  & 50 & 100 \\
     \midrule 
    \ourmethod{} (ours) & $90.94 \pm 2.18$ & $87.02 \pm 0.22$ & $85.3 \pm 1.81$\\
    \bottomrule
    \end{tabular}
    %}
    \label{tab:specnorm}
\end{table}

 We show in Theorem 1 that the generalization is affected by the hypernetworks Lipschitz constant $L_h$. This theoretical result suggests that we can benefit from bounding this constant. Here we empirically test this by applying spectral normalization~\cite{spectral_normalization} for all layers of the HN. The results are presented in Table~\ref{tab:specnorm}. We do not observe any significant improvement compared to the results without spectral normalization (presented in Table 1 of the main text).
% \renewcommand{\tabcolsep}{3pt}
% \begin{table}[h]
%     \vskip 0.15in
%     \small
%     %\tiny
%     \centering
%     %\scalebox{0.9}{
%     \caption{\ourmethod{} with spectral-normalization.}
%     \begin{tabular}{l c c c }
%     \toprule
%     & \multicolumn{3}{c}{CIFAR10}\\
%      \cmidrule{2-4}\\
%      & 10  & 50 & 100 \\
%      \midrule 
%     \ourmethod{} (ours) & $90.94 \pm 2.18$ & $87.02 \pm 0.22$ & $85.3 \pm 1.81$\\
%     \bottomrule
%     \end{tabular}
%     %}
%     \label{tab:specnorm}
% \end{table}

% \subsection{Generalization}
% \begin{table*}[th]
%     \vskip 0.15in
%     % \small
%     \scriptsize
%     \centering
%     \caption{Generalization to novel clients: test accuracy for CIFAR10/100.}
%     \begin{tabular}{l c c c c c}
%     \toprule
%     & \multicolumn{2}{c}{CIFAR10} && \multicolumn{2}{c}{CIFAR100} \\
%      \cmidrule{2-3} \cmidrule{5-6} \\
%     & Training clients & Novel clients &&
%      Training clients & Novel clients\\
%     \midrule
%     Per-FedAvg & $80.60 \pm 0.49$ & $81.29 \pm 1.81$ && $47.37 \pm 0.81$ & $44.05 \pm 0.91$\\
%     pFedMe & $84.38 \pm 0.19$ & $82.09 \pm 1.84$ && $46.53 \pm 0.84$ & $45.19 \pm 0.54$\\
%     \midrule
%     \ourmethod{}-PC (ours) & $88.22 \pm .65$ & $88.45 \pm 1.49$ && $50.68 \pm 1.43$ & $49.56 \pm 2.19$\\
%     \bottomrule
%     \end{tabular}
%     \label{tab:gen}
% \end{table*}

\subsection{Generalization to Novel Clients}

\begin{figure}[h]
    \centering
    \includegraphics[width=1.\linewidth]{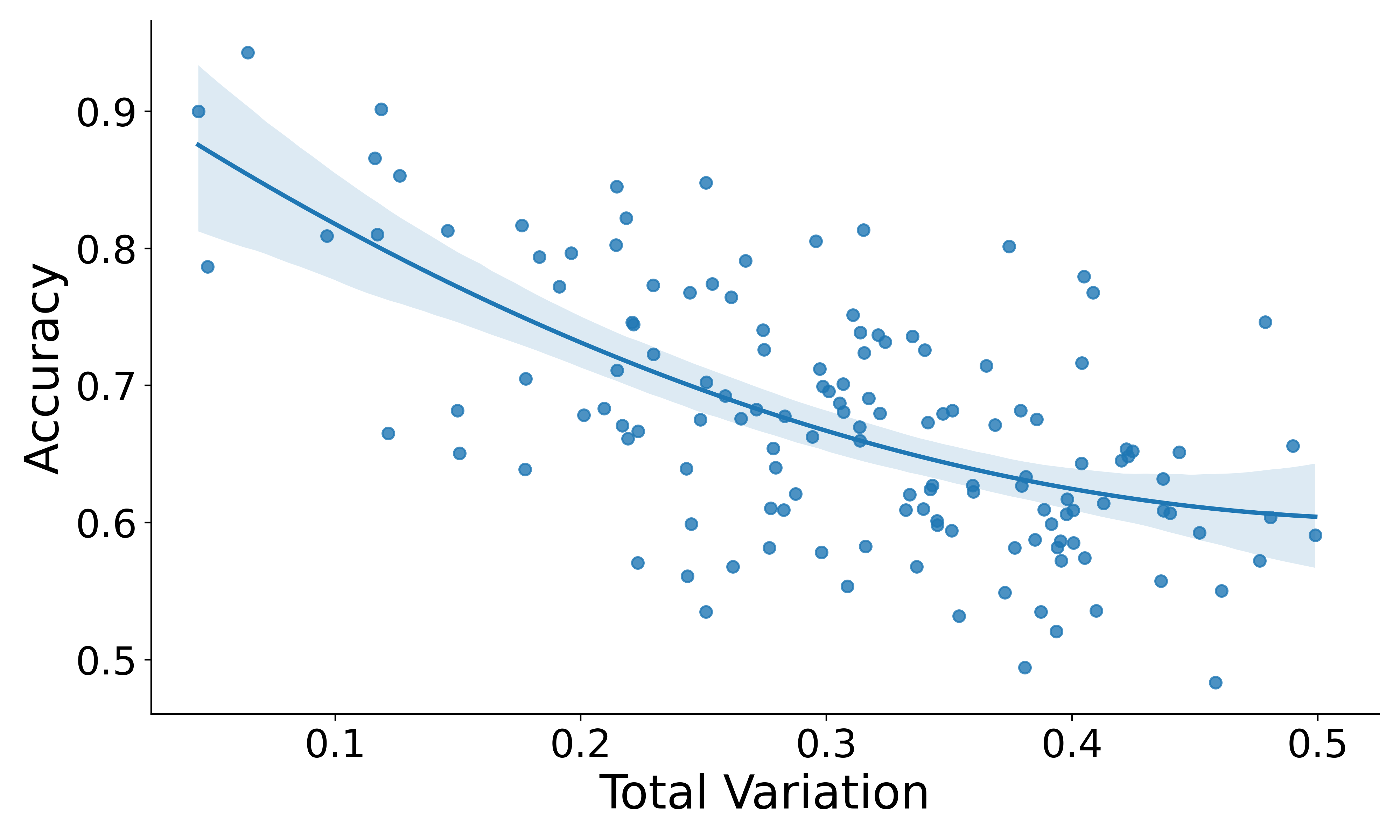}
    \caption{Accuracy for novel clients on the CIFAR10 test set. Each point represents a different client. Total variation is computed w.r.t the nearest training set client.}
    \label{fig:gen_raw}
\end{figure}

Here we provide additional results on the generalization performance for novel clients, studied in Section 5.3 of the main text. Figure~\ref{fig:gen_raw} shows the accuracy of individual clients as a function of the total variation distance. Each point represents a different client, where the total variation distance is calculated w.r.t to the nearest training set client. As expected, the results show (on average) that the test accuracy decreases with the increase in the total variation distance.

\subsection{Fixed Client Representation}

We wish to compare the performance of \ourmethod{} when trained with a fixed vs trainable client embedding vectors. We use CIFAR10 with the data split described in Section 5.3 of the main text and $50$ clients. We use a client embedding dimension of $10$. We set the fixed embedding vector for client $i$ to the vector of class proportions, $\hat{p}_i$, described in Section 5.3. \ourmethod{} achieves similar performance with both the trainable and fixed client embedding, $84.12\pm0.42$ and $83.92\pm0.36$ respectively.

\end{document}